\documentclass[runningheads]{llncs}
\usepackage{graphicx}
\usepackage{tikz}
\usepackage{comment}
\usepackage{amsmath,amssymb} 
\usepackage{color}

\usepackage{wrapfig}
\usepackage{algorithm}
\usepackage{algorithmic}
\usepackage{amsfonts}
\usepackage{subfigure}
\usepackage{booktabs}
\usepackage{multirow} 
\usepackage{colortbl}
\usepackage{wrapfig}
\usepackage{hyperref}
\hypersetup{
  colorlinks=true,
  linkcolor=red,
  filecolor=red,
  urlcolor=red,
  citecolor=cyan,
}

\makeatletter 
  \newcommand\figcaption{\def\@captype{figure}\caption} 
  \newcommand\tabcaption{\def\@captype{table}\caption} 
\makeatother

\usepackage[accsupp]{axessibility}  

\usepackage[capitalize]{cleveref}
\crefname{section}{Sec.}{Secs.}
\Crefname{section}{Section}{Sections}
\Crefname{table}{Table}{Tables}
\crefname{table}{Tab.}{Tabs.}

\begin{document}
\pagestyle{headings}
\mainmatter
\def\ECCVSubNumber{6519}  

\title{Tree Structure-Aware Few-Shot Image Classification via Hierarchical Aggregation} 

\titlerunning{Few-Shot Image Classification via Hierarchical Aggregation}
\author{Min Zhang$^{124}$ \quad Siteng Huang$^{24}$ \quad Wenbin Li$^{3}$ \quad Donglin Wang$^{24}$\thanks{Corresponding author}}
\institute{$^{1}$Zhejiang University \quad $^{2}$Westlake University \\
$^{3}$State Key Laboratory for Novel Software Technology, Nanjing University \\ 
$^{4}$Institute of Advanced Technology, Westlake Institute for Advanced Study \\
\email{liwenbin@nju.edu.cn, \{zhangmin,huangsiteng,wangdonglin\}@westlake.edu.cn}} 
\authorrunning{M. Zhang et al.}

\maketitle

\begin{abstract}
    In this paper, we mainly focus on the problem of how to learn additional feature representations for few-shot image classification through pretext tasks (\textit{e.g.}, rotation or color permutation and so on). This additional knowledge generated by pretext tasks can further improve the performance of few-shot learning (FSL) as it differs from human-annotated supervision (\textit{i.e.}, class labels of FSL tasks). To solve this problem, we present a plug-in \textit{Hierarchical Tree Structure-aware (HTS)} method, which not only learns the relationship of FSL and pretext tasks, but more importantly, can adaptively select and aggregate feature representations generated by pretext tasks to maximize the performance of FSL tasks. A hierarchical tree constructing component and a gated selection aggregating component is introduced to construct the tree structure and find richer transferable knowledge that can rapidly adapt to novel classes with a few labeled images. Extensive experiments show that our HTS can significantly enhance multiple few-shot methods to achieve new state-of-the-art performance on four benchmark datasets. The code is available at: \url{https://github.com/remiMZ/HTS-ECCV22}.
\end{abstract}
\section{Introduction}
\vspace{-5pt}
Few-shot learning (FSL), especially few-shot image classification~\cite{Fei-FeiFP03,TianWKTI20,BateniGMWS20,LiuHLJL21,zhang2022domain}, has attracted a lot of machine learning community. FSL aims to learn transferable feature representations by training the model with a collection of FSL tasks on base (seen) classes and generalizing the representations to novel (unseen) classes by accessing an extremely few labeled images~\cite{AnXZZ21,CuiG21,vinyals2016,FinnAL17,Requeima0BNT19,ChenLKWH19}. However, due to the data scarcity, the learned supervised representations mainly focus on the differences between the base class while ignoring the valuable semantic features within images for novel classes, weakening the model's generalization ability. Therefore, more feature representations should be extracted from the limited available images to improve the generalization ability of the FSL model.

\begin{figure}[htbp]
  \vspace{-2pt}
  \centering
  \includegraphics[width=0.65\linewidth]{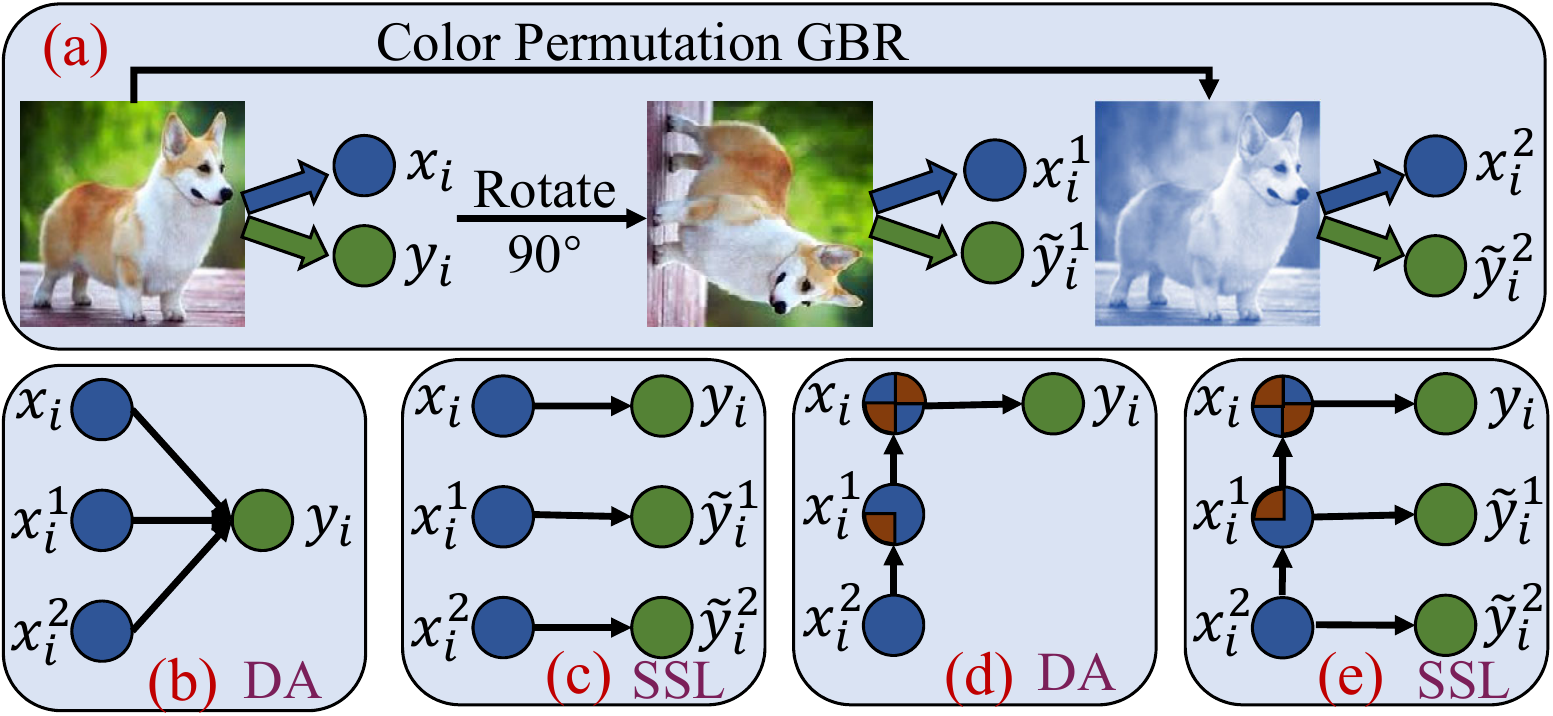}
  \vspace{-8pt}
	\caption{Differences in the learning process for few-shot image classification using pretext tasks between previous and our works. (a) shows the process of generating augmented images using FSL images. (b) and (c) show the learning process for previous works under the DA or SSL setting, which uses all images indiscriminately. (d) and (e) show the learning process for our work under the DA or SSL setting, which can exploit the hierarchical tree structure to adaptively select useful feature representations.} 
	\label{fig:intro}
  \vspace{-16pt}
\end{figure}

One effective way of extracting more useful feature representations is to use pretext tasks, such as rotation with multiple angles or color permutation among different channels~\cite{GidarisBKPC19,lee2020self,SuMH20,liu2019self,chen2021pareto}. Because these pretext tasks can generate additional augmented images and the semantic representations of these augmented images are a good supplementary to the normal human-annotated supervisions (\textit{i.e.,} class labels of FSL images), which is beneficial to the few-shot learning model generalization on novel classes. The standard training process of using pretext tasks to assist in FSL can be roughly classified into two settings, \textit{i.e.,} data augmentation (DA) and self-supervised learning (SSL), following existing works~\cite{GidarisBKPC19,SuMH20,NiGSKG21,liu2019self,TianWKTI20}.
As shown in~\cref{fig:intro}, (a) shows that pretext tasks are used to generate multiple augmented images ($x_{i}^{1}$, $x_{i}^{2}$) and $x_{i}$ is FSL images. (b) and (c) are the learning process of using pretext tasks to improve the FSL performance under the DA or SSL setting in previous works. However, in the DA setting, all augmented and raw images are placed in the same label space (\textit{e.g.}, $y_{i}=\tilde{y}_{i}^{1}=\tilde{y}_{i}^{2}=$dog) and the empirical risk minimization (ERM) (see~\cref{Eq:DA}) is used to optimize the model, making the model use all images indiscriminately.

We find that when augmented images are generated by using inappropriate pretext tasks, this optimization method (\textit{i.e.}, considering the information of all images on average) may destroy the performance of the FSL task (see~\cref{fig:base}). This is because the augmented images bring ambiguous semantic information (\textit{e.g}, rotation for symmetrical object)~\cite{FengXT19,CubukZMVL19,MisraM20}. Although it can be solved using expert experience to select appropriate pretext tasks for different datasets, which is very labor consuming~\cite{NiGSKG21}. To this end, we believe that it is very important that the model can adaptively select augmented image features to improve the performance of the FSL task.
In the SSL setting (\cref{fig:intro} (c)), it retains FSL as the main task and uses pretext tasks as additional auxiliary tasks (\textit{i.e.,} SSL tasks). From~\cref{fig:base}, we find that SSL using independent label spaces (\textit{e.g.}, $y_{i}$=dog, $\tilde{y}_{i}^{1}$=$90^{\circ}$, $\tilde{y}_{i}^{2}$=GBR) to learn these tasks separately (see~\cref{Eq:SSL}) can alleviate the problem caused by DA training based on one single label space, but it is not enough to fully learn the knowledge hidden in these augmented images only through the sharing network.
This is because there are similarities among augmented images generated under the same raw image and different pretext tasks, and the relationships among these augmented images should be modeled.

To effectively learn knowledge from augmented images, we propose a plug-in \textit{Hierarchical Tree Structure-aware (HTS)} method for few-shot classification. The core of the proposed HTS method: (1) using a tree structure to model the relationships of raw and augmented images; (2) using a gated aggregator to adaptively select the feature representations to improve the performance of the FSL task. Next, we outline two key components of the proposed HTS method.

\textbf{Modeling the relationships via a hierarchical tree constructing component.}
This component aims to construct a tree structure for each raw image, and thus we can use the edge of the tree to connect the feature information among different augmented images and use the level of the tree to learn the feature representations from different pretext tasks. In addition, when pretext tasks or augmented images change (\textit{e.g.}, adding, deleting or modifying), it is very flexible for our HTS method to change the number of levels or nodes.

\textbf{Adaptively learning features via a gated selection aggregating component.}
In this paper, we use \textit{Tree-based Long Short Term Memory (TreeLSTM)}~\cite{TaiSM15} as the gated aggregator following the reasons as below:
(1) On the above trees, we find that the augmented images (\textit{i.e.}, nodes) from different pretext tasks can be further formulated as a sequence with variable lengths from the bottom to the top level.
(2) TreeLSTM generates a forgetting gate for each child node, which is used to filter the information of the corresponding child nodes (different colors are shown in~\cref{fig:intro} (d) and (e)).
This indicates that the representations of the lower-level nodes can be sequentially aggregated and enhance the upper-level nodes' outputs.
Finally, these aggregated representations will be used in training and testing phases. The main contributions of HTS are: 

1. We point out the limitations of using pretext tasks to help the few-shot model learn richer and transferable feature representations. To solve these limitations, we propose a hierarchical tree structure-aware method.

2. We propose a hierarchical tree constructing component to model the relationships of augmented and raw images and a gated selection aggregating component to adaptively learn and improve the performance of the FSL task.

3. Extensive experiments on four benchmark datasets demonstrate that the proposed HTS is significantly superior to the state-of-the-art FSL methods.
\section{Related Work}
\label{sec:rela}

\subsection{Few-Shot Learning}
The recent few-shot learning works are dominated by meta-learning based methods. They can be roughly categorized into two groups:
(1) \textit{Optimization-based methods} advocate learning a suitable initialization of model parameters from base classes and transferring these parameters to novel classes in a few gradient steps~\cite{RusuRSVPOH19,ZhangKD20,BertinettoHTV19,FinnAL17,QiaoLSY18}. 
(2) \textit{Metric-based methods} learn to exploit the feature similarities by embedding all images into a common metric space and using well-designed nearest neighbor classifiers~\cite{li2019revisiting,li2019distribution,kang2021relational,ZhangCLS20}. In this paper, our HTS can equip with an arbitrary meta-learning based method and improve performance.

\vspace{-2pt}
\subsection{Pretext Tasks}
\vspace{-2pt}
Pretext tasks have succeeded to learn useful representations by focusing on richer semantic information of images to significantly improve the performance of image classification. In this paper, we are mainly concerned with the works of using pretext tasks to improve the performance of few-shot classification~\cite{GidarisBKPC19,gidaris2018rotation,zhang2020iept,lee2020self,chen2021pareto}. However, these works are often shallow, \textit{e.g.}, the original FSL training pipeline is intact and an additional loss (self-supervised loss) on each image is introduced, leading to the learning process not being able to fully exploit the augmented image representations. Different from these works, we introduce a hierarchical tree structure (HTS) to learn the pretext tasks. Specifically, the relationships of each image are modeled to learn more semantic knowledge. Moreover, HTS can adaptively select augmented features to avoid the interference of ambiguous information. Our experimental results show that thanks to the reasonable learning of pretext tasks, our HTS method clearly outperforms these works (see \cref{Table:SOTA}).
\vspace{-16pt}
\section{Preliminaries}
\label{sec:pre}

\vspace{-2pt}
\subsection{Problem Setting in Few-Shot Learning}
\label{subsec:fsl}
\vspace{-5pt}
We consider that meta-learning based methods are used to solve the few-shot classification problem, and thus follow the episodic (or task) training paradigm. 
In the meta-training phase, we randomly sample episodes from a base class set $\mathcal{D}_{b}$ to imitate the meta-testing phase sampled episodes from a novel class set $\mathcal{D}_{n}$.
Note that $\mathcal{D}_{b}$ contains a large number of labeled images and classes but has a \textit{disjoint} label space with $\mathcal{D}_{n}$ (\textit{i.e.} $\mathcal{D}_{b} \cap \mathcal{D}_{n} = \varnothing$ ).
Each $n$-way $k$-shot episode $\mathcal{T}_{e}$ contains a support set $\mathcal{S}_{e}$ and a query set $\mathcal{Q}_{e}$.
Concretely, we first randomly sample a set of $n$ classes $\mathcal{C}_{e}$ from $\mathcal{D}_{b}$, and then generate $\mathcal{S}_{e} = \{(x_{i},y_{i})|y_{i}\in \mathcal{C}_{e}, i=1, \cdots, n\times k\}$ and $\mathcal{Q}_{e} = \{(x_{i},y_{i})|y_{i}\in \mathcal{C}_{e}, i=1, \cdots, n\times q\}$ by sampling $k$ support and $q$ query images from each class in $\mathcal{C}_{e}$, and $\mathcal{S}_{e} \cap \mathcal{Q}_{e} = \varnothing$. 
For simplicity, we denote $l_{k} = n\times k$ and $l_{q} = n\times q$.
In the meta-testing phase, the trained few-shot learning model is fine-tuned using the support set $\mathcal{S}_{e}$ and is tested using the query set $\mathcal{Q}_{e}$, where the two sets are sampled from the novel class set $\mathcal{D}_{n}$.

\vspace{-8pt}
\subsection{Few-Shot Learning Classifier}
\label{subsec:class}
\vspace{-5pt}
We employ ProtoNet~\cite{SnellSZ17} as the few-shot learning (FSL) model for the main instantiation of our HTS framework due to its simplicity and popularity. 
However, we also show that any meta-learning based FSL method can be combined with our proposed HTS method (see results in~\cref{Table:adapting}). 
ProtoNet contains a feature encoder $E_{\phi}$ with learnable parameters $\phi$ (\textit{e.g.}, CNN) and a simple non-parametric classifier.
In each episode $\mathcal{T}_{e} = \{\mathcal{S}_{e}, \mathcal{Q}_{e}\}$, ProtoNet computes the mean feature embedding of the support set for each class $c\in \mathcal{C}_{e}$ as the prototype $\tilde{p}_{c}$:
\vspace{-8pt}
\begin{equation} 
  \label{Eq:prototy}
  \begin{aligned}
    \tilde{p}_{c} = \frac{1}{k} \sum\nolimits_{(x_{i},y_{i}\in \mathcal{S}_{e})}E_{\phi}(x_{i})\cdot\mathbb{I}(y_{i} = c),
  \end{aligned}
  \vspace{-5pt}
\end{equation}
where $\mathbb{I}$ is the indicator function with its output being 1 if the input is true or 0 otherwise. 
Once the class prototypes are obtained from the support set, ProtoNet computes the distances between the feature embedding of each query set image and that of the corresponding prototypes.
The final loss function over each episode using the empirical risk minimization (ERM) is defined as follows:
\vspace{-6pt}
\begin{equation} 
  \label{Eq:fsl} 
  \begin{aligned}
    &L_{FSL}(\mathcal{S}_{e}, \mathcal{Q}_{e}) = \frac{1}{|\mathcal{Q}_{e}|}\sum\nolimits_{(x_{i},y_{i}\in \mathcal{Q}_{e})} -log \ p_{y_{i}}, \\
    &p_{y_{i}} = \frac{exp(-d(E_{\phi}(x_{i}),\tilde{p}_{y_{i}}))}{\sum\nolimits_{c\in \mathcal{C}_{e}}exp(-d(E_{\phi}(x_{i}),\tilde{p}_{c}))},
  \end{aligned}
  \vspace{-5pt}
\end{equation}
where $d(\cdot, \cdot)$ denotes a distance function (\textit{e.g.}, the squared euclidean distance for ProtoNet method following the original paper~\cite{SnellSZ17}).
\vspace{-6pt}
\section{Methodology}
\label{sec:meth}

\vspace{-3pt}
\subsection{Pretext Tasks in FSL}
\label{subsec:pretext}
\vspace{-5pt}
Pretext tasks assisting in few-shot learning have two settings: data augmentation (DA) and self-supervised learning (SSL) (see the schematic in~\cref{fig:intro}).
We first define a set of pretext-task operators $\mathcal{G} = \{g_{j}|j = 1,\cdots, J\}$, where $g_{j}$ means the operator of using the $j$-th pretext task and $J$ is the total number of pretext tasks. 
Moreover, we also use $M_{j}$ to represent the number of augmented images generated by using the $j$-th pretext task for each raw image and the pseudo label set of this task is defined as $\tilde{Y}^{j} = \{0,\cdots, M_{j}-1\}$.
For example, for a 2D-rotation operator, each raw image will be rotated with multiples of 90$^\circ$ angles (\textit{e.g.}, 90$^\circ$, 180$^\circ$, 270$^\circ$), where the augmented images are $M_{rotation} = 3$ and the pseudo label set is $\tilde{Y}^{rotation} = \{0, 1, 2\}$. 
Given a raw episode $\mathcal{T}_{e} = \{\mathcal{S}_{e},\mathcal{Q}_{e}\}$ as described in~\cref{subsec:fsl}, we utilize these pretext-task operators from $\mathcal{G}$ in turn to augment each image in $\mathcal{T}_{e}$.
This results in a set of $J$ augmented episodes are 
$\mathcal{T}_{aug} = \{(x_{i},y_{i},\tilde{y}_{i},j)|y_{i}\in \mathcal{C}_{e}, \tilde{y}_{i}\in \tilde{Y}^{j}, i=1,\cdots, M_{j}\times l_{k}, M_{j}\times (l_{k}+1), \cdots, M_{j}\times (l_{k}+l_{q}), j=1, \cdots, J\}$, where the first images $M_{j}\times l_{k}$ are from the augmented support set $\mathcal{S}_{e}^{j}$ and the rest of images $M_{j}\times l_{q}$ from the augmented query set $\mathcal{Q}_{e}^{j}$. 

\noindent \textbf{Data Augmentation.}
For DA setting, we use the combined episodes $\mathcal{T} = \{\{\mathcal{S}_{e}^{r}, \mathcal{Q}_{e}^{r}\}|r=0,\cdots, J\}$, where $\{\mathcal{S}_{e}^{0}, \mathcal{Q}_{e}^{0}\}$ is the raw episode and $\{\{\mathcal{S}_{e}^{r}, \mathcal{Q}_{e}^{r}\}|r=1,\cdots, J\}$ is the augmented episodes.
In this paper, when $r\geq 1$, the value of $r$ is equal to $j$ unless otherwise stated. 
Each image $(x_{i}, y_{i})$ in $\mathcal{T}$ takes the same class label $y_{i}$ (from the human annotation) for supervised learning to imporve the performance of the FSL. 
The objective is to minimize a cross-entropy loss: 
\vspace{-8pt}
\begin{equation} \label{Eq:DA}
  \begin{aligned}
    L_{DA} = \frac{1}{J+1}\sum\nolimits_{r=0}^{J}L_{FSL}(\mathcal{S}_{e}^{r}, \mathcal{Q}_{e}^{r}).
  \end{aligned}
  \vspace{-7pt}
\end{equation}

$L_{DA}$ uses the empirical risk minimization (ERM) algorithm based on the same label space (\textit{e.g.}, $y_{i}$) to learn raw and augmented feature representations.
However, if the augmented images have ambiguous representations, this optimization method may interfere with the semantic learning of the FSL model.

\noindent \textbf{Self-Supervised Learning.} 
For SSL setting, each raw image $(x_{i}, y_{i})$ in $\mathcal{T}_{e}$ uses a class label $y_{i}$ for supervised learning, while each augmented image $(x_{i}, \tilde{y}_{i})$ in $\mathcal{T}_{aug}$ carries a pseudo label $\tilde{y}_{i}$ for self-supervised learning. 
A multi-task learning loss (FSL main task and SSL auxiliary task) is normally adopted as below:
\vspace{-8pt}
\begin{equation}
	\begin{aligned}
	  &L_{SSL}= L_{FSL}(\mathcal{S}_{e}, \mathcal{Q}_{e}) + \sum\nolimits_{j=1}^{J}\beta_{j}L_{j}, \nonumber
	\end{aligned}
  \vspace{-6pt}
\end{equation}

\vspace{-10pt}
\begin{equation}
  \label{Eq:SSL}
	\begin{aligned}
    L_{j} = &\frac{1}{E}\sum\nolimits_{(x_{i}, \tilde{y}_{i}\in \mathcal{T}_{e}^{j})} -log \frac{exp([\theta^{j}(E_{\phi}(x_{i}))]_{\tilde{y}_{i}})}{\sum\nolimits_{\tilde{y}^{'}}^{\tilde{Y}^{j}}exp([\theta^{j}(E_{\phi}(x_{i}))]_{\tilde{y}^{'}})}, 
	\end{aligned}
  \vspace{-3pt}
\end{equation}
where $E=M_{j}\times (l_{k}+l_{q})$, $[\theta^{j}(E_{\phi}(x_{i}))]$ denotes the $j$-th pretext task scoring vector and $[.]_{\tilde{y}}$ means taking the $\tilde{y}$-th element.
$L_{SSL}$ learns the knowledge of the few-shot learning and multiple sets of pretext tasks in different label spaces, but only uses a shared feature encoder $E_{\phi}$ to exchange these semantic information.

\vspace{-5pt}
\subsection{Pretext Tasks in HTS}
\label{subsec:HTS}
\vspace{-5pt}
In this paper, we propose a hierarchical tree structure-aware (HTS) method that uses a tree to model relationships and adaptively selects knowledge among different image features. 
There are two key components in HTS: hierarchical tree constructing component and gated selection aggregating component \footnote{Note that our method mainly focuses on how to adaptively learn the knowledge of pretext tasks and improve the performance of few-shot image classification.}.

\vspace{-10pt}
\subsubsection{Hierarchical Tree Constructing Component}
\label{subsubsec:hts} 
\ 
\vspace{2pt}
\newline
Given augmented episodes $\mathcal{T}_{aug}=\{\mathcal{T}_{e}^{j}=\{\mathcal{S}_{e}^{j}, \mathcal{Q}_{e}^{j}\}|j=1,\cdots, J\}$ as described in~\cref{subsec:pretext}, each augmented episode in $\mathcal{T}_{aug}$ corresponds to one specific pretext task of the same set of images from the raw episode $\mathcal{T}_{e}$.
Therefore, we believe that these augmented images with different pretext tasks should be modeled to capture the correlations and further learn more semantic feature representations. 

To this end, we construct a tree structure for each FSL image and its corresponding multiple sets of augmented images generated by using different pretext tasks in each episode.
Specifically, \textcolor{blue}{(1)} we extract the feature vectors of the raw and augmented episodes by using the shared feature encoder $E_{\phi}$ and denote the feature set as $\mathcal{T}_{emd}$, where $\mathcal{T}_{emd}=\{E_{\phi}(x_{i})|(x_{i},y_{i},\tilde{y}_{i}, r)\in \mathcal{T}_{e}^{r}, r=0,\cdots, J, i=1,\cdots, M_{j}\times (l_{k}+l_{q})\}$.
\textcolor{blue}{(2)} The feature vectors of the raw episode $\mathcal{T}_{emd}^{0}$ are taken as the root nodes and each raw image has its own tree structure.
\textcolor{blue}{(3)} The augmented feature vectors of these augmented episodes $\mathcal{T}_{emd}^{j}$ with the $j$-th pretext task are put in the $(j+1)$-th level of the tree.
\textcolor{blue}{(4)} We take a raw image $x_{i}$ (\textit{i.e}, a tree structure) and its multiple sets of augmented images $\{x_{i}^{j}\}$ as an example to indicate how to construct this tree, and repeat the process for other raw images.
The form of the tree is  
\{$E_{\phi}(x_{i})\overset{g_{1}}{\to}E_{\phi}(x_{i}^{1})\overset{g_{2}}{\to}\cdots\overset{g_{j}}{\to}E_{\phi}(x_{i}^{j})\cdots\overset{g_{J}}{\to}E_{\phi}(x_{i}^{J})$\}, where the raw feature set $E_{\phi}(x_{i})$ is in the $1$-st level (root nodes) and the augmented feature set is in the $(j+1)$-th level sharing the same pretext task $g_{j}$.
For each episode, we construct $(l_{k}+l_{q})$ hierarchical tree structures, and every level has $M_{j}$ child nodes with the $j$-th pretext task.
In these tree structures, the edge information is used to model the relationships of different augmented or raw images.    
The level knowledge is used to learn the representations from different pretext tasks.
In~\cref{subsubsec:gated}, we introduce how to better aggregate image features.

\begin{figure}[tbp]
	\centering 
  \vspace{-2pt}
  \includegraphics[width=0.6\linewidth]{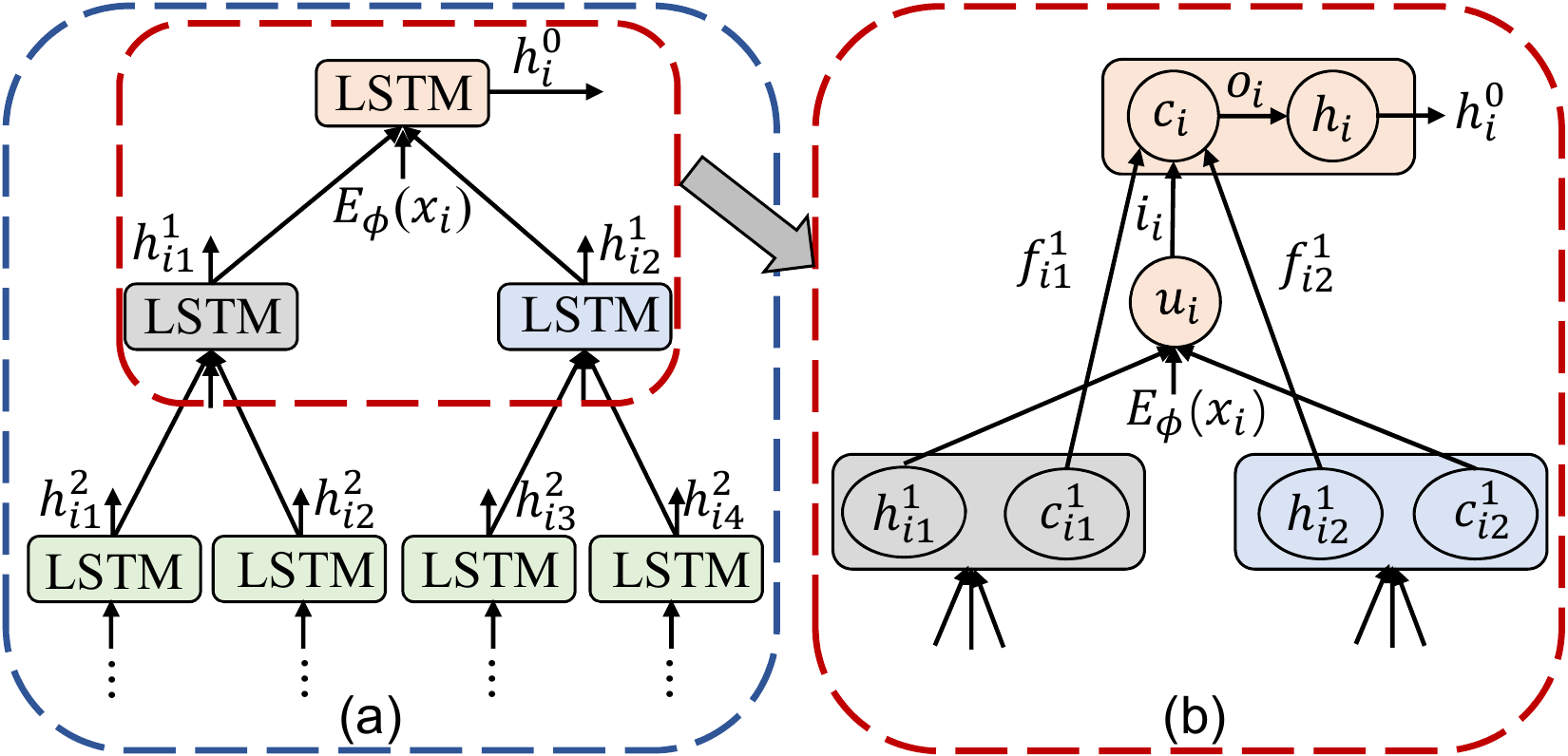}
  \vspace{-8pt}
	\caption{The learning process of gated selection aggregating component.
  (a) shows that the aggregator to sequentially and hierarchically aggregate the information from bottom to top level.
  (b) details the internal aggregation of TreeLSTM (\textit{e.g.}, two levels).
  The subscript marked in this figure represents the number of child nodes, the superscript represents the number of levels and different colors are different LSTM cells.} 
	\label{fig:treelstm}
  \vspace{-16pt} 
\end{figure}

\vspace{-10pt}
\subsubsection{Gated Selection Aggregating Component} 
\label{subsubsec:gated}
\
\vspace{2pt}
\newline
As mentioned above, we have constructed a tree structure for each raw image $x_{i}$ in each randomly sampled episode.
The following question is how to effectively use this tree structure for learning and inference.
Firstly, our intuition is to preserve the tree structure information, because it models the relationships among images (\textit{e.g.}, FSL and augmentations). 
Secondly, we should selectively aggregate the features of all child nodes from the bottom to the top level, because the information aggregated process aims to maximize the performance of the parent node after aggregation.
Thirdly, since the feature information in the hierarchical propagation can be regarded as sequential inputs divided by levels but meta-learning can not directly process the sequential data.
Finally, we adopt the \textit{tree-based Long Short Term Memory (TreeLSTM)} as our gated aggregator to encode the lower-level information into an upper-level output. 
In this way, we can mine much richer features from the tree structure (see~\cref{fig:treelstm}).
Next, we will detail the aggregation and propagation process of TreeLSTM on these trees.

For simplicity, we take a tree structure as an example to introduce how to selectively aggregate information by using TreeLSTM aggregator and repeat the process for other tree structures. 
The $(J+1)$-level tree structure are constructed based on raw image features $E_{\phi}(x_{i})$ and its multiple sets of augmented features $\{E_{\phi}(x_{i}^{j})|i=1,\cdots, M_{j}, j=1,\cdots, J\}$.
The TreeLSTM aggregates information step by step from bottom level (\textit{i.e}, $(J+1)$-th level) to top level (\textit{i.e.}, $1$-st or root node level).
We use $\{h_{i}^{r}|r=0,\cdots, J\}$ to represent the aggregated node representations of each level except the bottom-level nodes in this tree.
Because the bottom node has no child nodes, its aggregate information is itself $\{E_{\phi}(x_{i}^{J})\}$.
The aggregation process can be formalized as: \{$h_{i}^{0}\overset{agg}{\longleftarrow}h_{i}^{1}\overset{agg}{\longleftarrow}\cdots\overset{agg}{\longleftarrow}h_{i}^{r}\cdots\overset{agg}{\longleftarrow}E_{\phi}(x_{i}^{J})$\}, where $agg$ denotes the aggregation operation by using the TreeLSTM.
The aggregated output $h_{i}^{r}$ of each level is represented as:
\vspace{-5pt}
\begin{equation} 
	\begin{aligned}
	  h_{i}^{r} = TreeLSTM(s_{i}, \{h_{m}\}), m\in \mathcal{M}_{i}, 
	\end{aligned}
  \vspace{-5pt}
\end{equation}
where $s_{i}\in \{h_{i}^{r}\}$ is any node in the tree and $\mathcal{M}_{i}$ is the set of child nodes of the $i$-th node in $(r+1)$ level of the tree. 

Since this form of child-sum in TreeLSTM conditions its components on the sum of child hidden states $h_{m}$, it is a permutation invariant function and is well-suited for trees whose children are unordered.
But, we find that using the child-mean replaces the child-sum in our implementation for better normalization, compared with the original paper~\cite{TaiSM15}.
The formulation of the TreeLSTM is: 
\vspace{-6pt}
\begin{equation}
  \begin{aligned}
    f_{m} = \sigma (W_{f}s_{i}+U_{f}&h_{m}+b_{f}), \\
    h_{me} = \frac{\sum\nolimits_{m\in \mathcal{M}_{i}} h_{m}}{|\mathcal{M}_{i}|}, \ \ 
    u_{i} = tanh &(W_{u}s_{i}+U_{u}h_{me}+b_{u}),  \\
    o_{i} = \sigma (W_{o}s_{i}+U_{o}h_{me}+b_{o}), \ \ 
    i_{i} = &\sigma (W_{i}s_{i}+U_{i}h_{me}+b_{i}),  \\
    c_{i} = i_{i} \odot u_{i} + \frac{\sum\nolimits_{m\in \mathcal{M}_{i}} f_{m} \odot c_{m}}{|\mathcal{M}_{i}|}&, \ \  
    h_{i}^{r} = o_{i} \odot tanh(c_{i}), 
  \end{aligned}
  \vspace{-6pt}
  \label{Eq:treelstm}
\end{equation}
where $\odot$ denotes the element-wise multiplication, $\sigma$ is sigmoid function, $W_{a}, U_{a}, b_{a}$, $a\in\{i,o,f,u\}$ are trainable parameters, $c_{i}$, $c_{k}$ are memory cells, $h_{i}$, $h_{m}$ are hidden states and $h_{me}$ denotes the child nodes mean in $(r+1)$-th level.
From~\cref{Eq:treelstm} and~\cref{fig:treelstm}, the aggregator learns a forgetting gate for each child and uses it to adaptively select the child features that are beneficial to its parent node based on the trained parameters.
Finally, the aggregated episodes are denoted as $\mathcal{T}_{agg}=\{\{\mathcal{S}_{agg}^{r},\mathcal{Q}_{agg}^{r}\}|r=0,\cdots, J\}$, where $\mathcal{S}_{agg}^{r}=\{h_{i}^{r}|i=1,\cdots, M_{r}*l_{k})\}$ and $\mathcal{Q}_{agg}^{r}=\{h_{i}^{r}|i=1,\cdots, M_{r}*l_{q})\}$ with $M_{0}=1, M_{r}=M_{j}$ when $r\geq 1$.

\vspace{-2pt}
\subsection{Meta-Training Phase with HTS} 
\label{subsec:training}
\vspace{-2pt}
The meta-training models with HTS use previous settings (DA or SSL) to train the FSL model.
Given the aggregated episodes $\mathcal{T}_{agg}=\{\{\mathcal{S}_{agg}^{r},\mathcal{Q}_{agg}^{r}\}|r=0,\cdots, J\}$, where $\{\mathcal{S}_{agg}^{0}, \mathcal{Q}_{agg}^{0}\}$ is the raw representations aggregated by using its all augmented images (\textit{i.e.}, from $(J+1)$-th to $2$-nd of the tree) and $\{\mathcal{S}_{agg}^{r}, \mathcal{Q}_{agg}^{r}\}_{r=1}^{J}$ is the augmented representations with each pretext task aggregated by using other augmented images (\textit{e.g}, from $(J+1)$-th to $r$-th of the tree). 

\noindent \textbf{Data Augmentation.} 
For data augmentation (DA) setting, we only use the aggregated root node (raw images) $\{\mathcal{S}_{agg}^{0}, \mathcal{Q}_{agg}^{0}\}$ to train the FSL model.  
The cross-entropy loss can be defined as:
\vspace{-5pt}
\begin{equation} \label{Eq:hts-da} 
  \begin{aligned}
    L^{HTS}_{DA} = L_{FSL}(\mathcal{S}_{agg}^{0}, \mathcal{Q}_{agg}^{0}).
  \end{aligned}
  \vspace{-5pt}
\end{equation}
$L_{HTS}^{DA}$ uses the aggregated representations with more knowledge to train the few-shot image classification model, different from~\cref{Eq:DA}.

\noindent \textbf{Self-Supervised Learning.}
For SSL setting, the aggregated root nodes train the FSL main task and each aggregated augmented node $(h_{i}^{r}, \tilde{y}_{i})$ in $\{\mathcal{T}_{agg}^{r}=\{\mathcal{S}_{agg}^{r},\mathcal{Q}_{agg}^{r}\}|r=1,\cdots, J\}$ trains the SSL auxiliary task. 
With the pseudo label $\tilde{y}_{i}$ and every level aggregated features in the tree, the multi-task learning loss is:
\vspace{-5pt}
\begin{equation} \label{Eq:hts-ssl}
	\begin{aligned}
	  &L^{HTS}_{SSL} = L_{FSL}(\mathcal{S}_{agg}^{0}, \mathcal{Q}_{agg}^{0}) +\sum\nolimits_{r=1}^{J} \beta_{r}L_{r}, \\
    L_{r} = &\frac{1}{E}\sum\nolimits_{(h_{i}^{r},\tilde{y}_{i}\in \mathcal{T}_{agg}^{r})} -log \frac{exp([\theta^{r}(h_{i}^{r}))]_{\tilde{y}_{i}})}{\sum\nolimits_{\tilde{y}^{'}}^{\tilde{Y}^{j}}exp([\theta^{r}(h_{i}^{r})]_{\tilde{y}^{'}})}, 
	\end{aligned}
  \vspace{-2pt}
\end{equation}
where $E$ and $[\theta^{r}(\cdot)]_{\tilde{y}}$ have the same meanings as \cref{Eq:SSL}. 
and when $r\geq 1$, $\theta_{r} = \theta_{j}$.
For easy reproduction, we summarize the overall algorithm for FSL with HTS in Algorithm 1 in~\cref{asec:alg}.
Notably, our HTS is a lightweight method, consisting of a parameter-free tree constructing component and a simple gated aggregator component. 
Therefore, as a plug-in method, HTS will not introduce too much additional computing overhead \footnote{During training, for a 5-way 1-/5-shot setting, one episode time is 0.45/0.54s (0.39/0.50s for baseline) with 75 query images over 500 randomly sampled episodes.}.
Once trained, with the learned network parameters, we can perform the testing over the test episodes.

\vspace{-2pt}
\subsection{Meta-Testing Phase with HTS}
\label{subsec:testing}
\vspace{-2pt}
During the meta-testing stage, we find that using pretext tasks for both the query set and support set, and then aggregating features based on the trained aggregator can further bring gratifying performance with only a few raw images (see~\cref{fig:t-sne}). Therefore, the predicted label for $x_{i}\in \mathcal{Q}_{e}$ can be computed with~\cref{Eq:prototy} and aggregated raw features $h_{i}^{0}$ as:
\vspace{-5pt}
\begin{equation} \label{Eq:test}
	\begin{aligned}
	  y_{i}^{pred} = \mathop{argmax}_{y\in\mathcal{C}_{e}} \frac{exp(-d(h_{i}^{0},\tilde{p}_{y_{i}}))}{\sum\nolimits_{c\in \mathcal{C}_{e}}exp(-d(h_{i}^{0},\tilde{p}_{c}))}. 
	\end{aligned}
  \vspace{-5pt}
\end{equation}

\subsection{Connections to Prior Works}
\textbf{Hierarchical few-shot learning.}
Recently, some works have been proposed to solve the problem of few-shot learning using the hierarchical structure.
One representative work aims to improve the effectiveness of meta-learning by hierarchically clustering different tasks based on the level of model parameter~\cite{YaoWHL19}, while our method is based on the feature level with less running time and computation.
Another work~\cite{LiuZLJZ19} learns the relationship between fine-grained and coarse-grained images through hierarchical modeling, but it requires the data itself to be hierarchical, while our method can adapt to any dataset.

\noindent\textbf{Long short term memory few-shot learning.}
Some works~\cite{vinyals2016,YeHZS20,LiZCL17} use chain-based long short-term memory (ChainLSTM) to learn the relationship of images. However, our work uses tree-based long short-term memory (TreeLSTM) to learn structured features and meanwhile preserve tree-structured information.
\section{Experimental Results} 
\label{sec:exper}

\subsection{Experimental Setup}
\label{subsec:setup}

\noindent \textbf{Research Questions.} 
Research questions guiding the remainder of the paper are as follows: 
\textbf{RQ1.} What is the real performance of the heuristic combination of FSL main task and pretext auxiliary tasks?
\textbf{RQ2.} How effective is the proposed HTS framework for the FSL image classification task in both single-domain and cross-domain settings?
\textbf{RQ3.} Could the proposed HTS framework adaptively select augmented features for better improving the performance of the FSL main task?
\textbf{RQ4.} 
How does the proposed HTS method work (ablation study)? 

\noindent \textbf{Benchmark Datasets.}
All experiments are conducted on four FSL benchmark datasets, \textit{i.e.}, \emph{mini}ImageNet~\cite{vinyals2016}, \emph{tiered}ImageNet~\cite{RenTRSSTLZ18}, CUB-200-2011~\cite{WahCUB_200_2011} and CIFAR-FS~\cite{BertinettoHTV19}.
The \textit{mini}ImageNet and \textit{tiered}ImageNet are the subsets of the ILSVRC-12 dataset~\cite{DengDSLL009}. 
CUB-200-2011 is initially designed for fine-grained classification.
The resolution of all images in the three datasets is resized to $84\times 84$.
CIFAR-FS is a subset of CIFAR-100 and each image is resized to $32\times 32$. 
See~\cref{asec:set} for more details on the four few-shot benchmark datasets. 

\noindent \textbf{Implementation Details.}
We adopt the episodic training procedure under 5-way 1-shot and 5-shot settings~\cite{vinyals2016}.  
In each episode, 15 unlabeled query images per class are used for the training and testing phases.
We apply Conv4 (filters:64) and ResNet12 (filters: [64, 160, 320, 640]) as the encoder.
Our model is trained from scratch and uses the Adam optimizer with an initial learning rate $10^{-3}$. 
The hyperparameter $\beta_{j} = 0.1$ for all experiments.
Each mini-batch contains four episodes and we use a validation set to select the best-trained model.
For all methods, we train 60,000 episodes for 1-shot and 40,000 episodes for 5-shot~\cite{ChenLKWH19}. 
We use PyTorch with one NVIDIA Tesla V100 GPU to implement all experiments and report the \textbf{\emph{average accuracy} (\%)} with 95\% \emph{confidence interval} over the 10,000 randomly sampled testing episodes. 
For our proposed HTS framework, we consider ProtoNet as our FSL model unless otherwise stated, but note that our framework is broadly applicable to other meta-learning based FSL models. 
More implementation details are given in~\cref{asec:set}.

\noindent \textbf{Pretext Tasks.} 
Following~\cite{lee2020self}, as the entire input images during training is important for image classification, we choose two same pretext tasks: \emph{rotation} and \emph{color permutation}.
We also give some subsets of the two tasks to meet the needs of experiments.
The rotation task has \textbf{rotation1} (90$^{\circ}$), \textbf{rotation2} (90$^{\circ}$, 180$^{\circ}$), \textbf{rotation3} (90$^{\circ}$, 180$^{\circ}$, 270$^{\circ}$) and \textbf{rotation4} (0$^{\circ}$, 90$^{\circ}$, 180$^{\circ}$, 270$^{\circ}$).
The color permutation task has \textbf{color\_perm1} (GBR), \textbf{color\_perm2} (GBR, BRG), \textbf{color\_perm3} (RGB, GBR, BRG) and \textbf{color\_perm6} (RGB, GBR, BRG, GBR, GRB, BRG, BGR). 
Note that our method can use arbitrary pretext tasks as it adaptively learns efficient features and improves the performance of FSL task. 

\begin{figure}[tbp]
  \centering
	\includegraphics[width=0.4\textwidth]{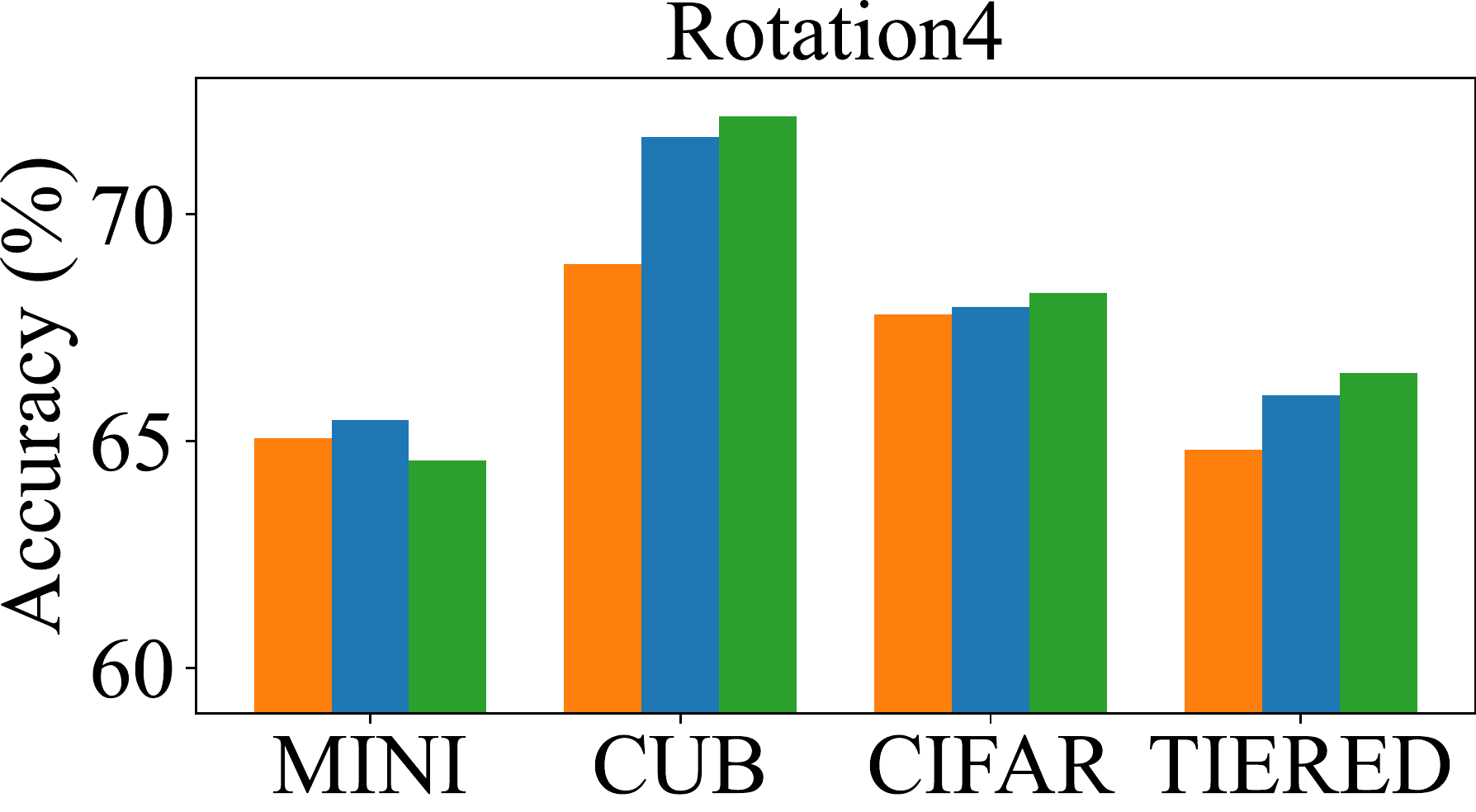}
  \label{fig:base-a}
  \includegraphics[width=0.4\textwidth]{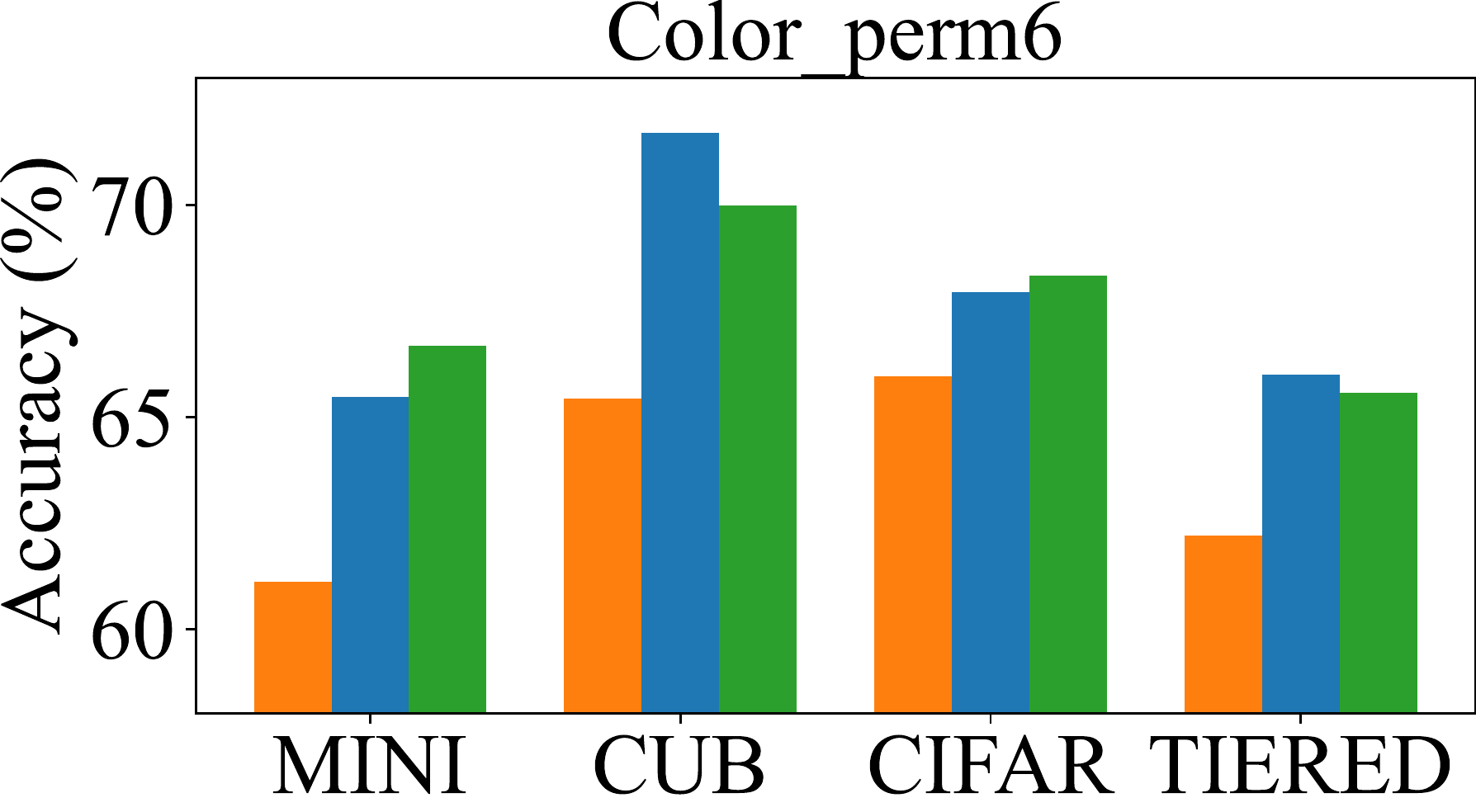}
  \label{fig:base-b} \\
  \includegraphics[width=0.5\textwidth]{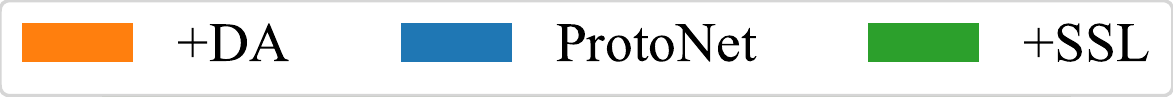}
  \vspace{-10pt}
  \caption{The results (5-shot) for motivation on \textit{mini}ImageNet (MINI), CUB-200-2011 (CUB), CIFAR-FS (CIFAR) and \textit{tiered}ImageNet (TIERED) with two pretext tasks.}
  \label{fig:base}
  \vspace{-15pt}
\end{figure}

\vspace{-6pt}
\subsection{RQ1. Performance of Pretext Tasks in FSL} 
\label{subsec:RQ1}
The experimental results in~\cref{fig:base} illustrate the key motivation of this paper and answer RQ1.
Intuitively, when we use the pretext tasks to generate multiple sets of augmented images, it can learn more knowledge and improve the performance of the FSL model, like~\cite{NiGSKG21}.
However, we find that not arbitrary pretext task has good performance on downstream datasets, and we need to find suitable tasks for different datasets.
The results of ProtoNet+DA are significantly lower than the baseline (ProtoNet), which proves our concern that it is unreasonable for empirical risk minimization (ERM) to treat all samples equally (see~\cref{Eq:DA}).   
For ProtoNet+SSL, the results are slightly higher than the baseline in most cases, which indicates that only using a shared encoder is not enough (see~\cref{Eq:SSL}).

\begin{table}[tbp]
  \vspace{-5pt}
	\newcommand{\tabincell}[2]{\begin{tabular}{@{}#1@{}}#2\end{tabular}}
	\centering
  \resizebox{0.65\linewidth}{!}{
	  \begin{tabular}{lccccccccccccc}
	  \toprule
	  Training $\rightarrow$Testing  &&& &\multicolumn{2}{c}{Method} & && &1-shot & & & &5-shot \\
	  \midrule
	  \multirow{2}[2]{*}{\tabincell{c}{\textit{mini}Imagenet$\rightarrow$ \\CUB-200-2011}} &&& & \multirow{2}[2]{*}{ProtoNet} & && & &32.18$\pm$\footnotesize0.25 & & & &55.95$\pm$\footnotesize0.21 \\
	  & & &&  &\cellcolor[rgb]{0.859, 0.859, 0.859}+HTS &\cellcolor[rgb]{0.859, 0.859, 0.859}&\cellcolor[rgb]{0.859, 0.859, 0.859}&\cellcolor[rgb]{0.859, 0.859, 0.859} & \cellcolor[rgb]{0.859, 0.859, 0.859}39.57$\pm$\footnotesize0.17 &\cellcolor[rgb]{0.859, 0.859, 0.859} &\cellcolor[rgb]{0.859, 0.859, 0.859} & \cellcolor[rgb]{0.859, 0.859, 0.859} &\cellcolor[rgb]{0.859, 0.859, 0.859}60.56$\pm$\footnotesize0.18 \\
    \midrule
	  \multirow{2}[2]{*}{\tabincell{c}{\textit{tiered}ImageNet$\rightarrow$ \\ CUB-200-2011}} & &&& \multirow{2}[2]{*}{ProtoNet} & && & &39.47$\pm$\footnotesize0.22 & & & &56.58$\pm$\footnotesize0.25 \\
	  & &&  &    & \cellcolor[rgb]{0.859, 0.859, 0.859}+HTS  &\cellcolor[rgb]{0.859, 0.859, 0.859} &\cellcolor[rgb]{0.859, 0.859, 0.859} &\cellcolor[rgb]{0.859, 0.859, 0.859}& \cellcolor[rgb]{0.859, 0.859, 0.859}42.24$\pm$\footnotesize0.20 &\cellcolor[rgb]{0.859, 0.859, 0.859} &\cellcolor[rgb]{0.859, 0.859, 0.859} &\cellcolor[rgb]{0.859, 0.859, 0.859} &\cellcolor[rgb]{0.859, 0.859, 0.859}60.71$\pm$\footnotesize0.18 \\
	  \midrule
	  \multirow{2}[2]{*}{\tabincell{c}{\textit{tiered}ImageNet$\rightarrow$ \\ \textit{mini}ImageNet}} 
	  & & && \multirow{2}[2]{*}{ProtoNet} & &&  & &47.01$\pm$\footnotesize0.26 & & & &66.82$\pm$\footnotesize0.25 \\
	  &   & &&   & \cellcolor[rgb]{0.859, 0.859, 0.859}+HTS & \cellcolor[rgb]{0.859, 0.859, 0.859} &\cellcolor[rgb]{0.859, 0.859, 0.859} &\cellcolor[rgb]{0.859, 0.859, 0.859}& \cellcolor[rgb]{0.859, 0.859, 0.859}55.29$\pm$\footnotesize0.20 &\cellcolor[rgb]{0.859, 0.859, 0.859} &\cellcolor[rgb]{0.859, 0.859, 0.859} &\cellcolor[rgb]{0.859, 0.859, 0.859} &\cellcolor[rgb]{0.859, 0.859, 0.859}72.67$\pm$\footnotesize0.16 \\
	  \bottomrule
	\end{tabular}}
  \vspace{1pt}
  \caption{Classification accuracy (\%) results comparison with 95\% confidence intervals for cross-domain evaluation with \textbf{rotation3}. More results see~\cref{asec:RQ2}.} 
\label{Table:cross-domain}
\vspace{-20pt}
\end{table}

\begin{table}[tbp]
  \begin{center}
    \resizebox{1.0\linewidth}{!}{
      \begin{tabular}{lcccccccccc}
        \toprule
        \multirow{2}[4]{*}{Method} & \multirow{2}[4]{*}{Backbone} & \multirow{2}[4]{*}{Tricks} & \multicolumn{2}{c}{\textbf{\emph{mini}ImageNet}} & \multicolumn{2}{c}{\textbf{CUB-200-2011}} & \multicolumn{2}{c}{\textbf{CIFAR-FS}} & \multicolumn{2}{c}{\textbf{\emph{tiered}ImageNet}} \\
        \cmidrule{4-11} &  &  & \multicolumn{1}{l}{1-shot} & \multicolumn{1}{l}{5-shot} & \multicolumn{1}{l}{1-shot} & \multicolumn{1}{l}{5-shot}& \multicolumn{1}{l}{1-shot} & \multicolumn{1}{l}{5-shot} & \multicolumn{1}{l}{1-shot} & \multicolumn{1}{l}{5-shot} \\
        \midrule
        MAML~\cite{FinnAL17}  & Conv4-32 & Fine-tuning & 48.70$\pm$\footnotesize1.84 & 55.31$\pm$\footnotesize0.73 &55.92$\pm$\footnotesize0.95 & 72.09$\pm$\footnotesize0.76 & 58.90$\pm$\footnotesize1.90 &71.50$\pm$\footnotesize1.00 &51.67$\pm$\footnotesize1.81 &70.30$\pm$\footnotesize1.75 \\
        PN~\cite{GidarisBKPC19} & Conv4-64 & Train-SSL & 53.63$\pm$\footnotesize0.43 &71.70$\pm$\footnotesize0.36 & - & - &64.69$\pm$\footnotesize0.32 &\textbf{80.82$\pm$\footnotesize0.24} &-& - \\
        CC~\cite{GidarisBKPC19} & Conv4-64 & Train-SSL &54.83$\pm$\footnotesize0.43 &71.86$\pm$\footnotesize0.33 & -&-&63.45$\pm$\footnotesize0.31 &79.79$\pm$\footnotesize0.24 & - & - \\
        closer \cite{ChenLKWH19}  & Conv4-64 & Train-DA  &48.24$\pm$\footnotesize0.75 &66.43$\pm$\footnotesize0.63 &60.53$\pm$\footnotesize0.83 &79.34$\pm$\footnotesize0.61 &-&-&-&-  \\
        CSS~\cite{AnXZZ21} & Conv4-64 & Train-SSL & 50.85$\pm$\footnotesize0.84 & 68.08$\pm$\footnotesize0.73 & 66.01$\pm$\footnotesize0.90 & \textbf{81.84$\pm$\footnotesize0.59} & 56.49$\pm$\footnotesize0.93 & 74.59$\pm$\footnotesize0.72 &- &- \\
        SLA~\cite{lee2020self} & Conv4-64 & Train-SSL & 44.95$\pm$\footnotesize0.79 &63.32$\pm$\footnotesize0.68 &48.43$\pm$\footnotesize0.82 &71.30$\pm$\footnotesize0.72 &45.94$\pm$\footnotesize0.87 &68.62$\pm$\footnotesize0.75 &- & -\\
        PSST~\cite{chen2021pareto} & Conv4-64 & Train-SSL & 57.04$\pm$\footnotesize0.51 & 73.17$\pm$\footnotesize0.48 & - & - & 64.37$\pm$\footnotesize0.33 & 80.42$\pm$\footnotesize0.32 & - & - \\
        CAN~\cite{HouCMSC19} & Conv4-64 & Train-DA & 52.04$\pm$\footnotesize0.00 & 65.54$\pm$\footnotesize0.00 & - &- & - & - & 52.45$\pm$\footnotesize0.00 & 66.81$\pm$\footnotesize0.00 \\
        \textbf{HTS (ours)}  & Conv4-64 & Train\&test-SSL  &\textbf{58.96$\pm$\footnotesize0.18} &\textbf{75.17$\pm$\footnotesize0.14} &\textbf{67.32$\pm$\footnotesize0.24} &78.09$\pm$\footnotesize0.15 &\textbf{64.71$\pm$\footnotesize0.21} &76.45$\pm$\footnotesize0.17 &\textbf{53.20$\pm$\footnotesize0.22} &\textbf{72.38$\pm$\footnotesize0.19} \\
        \midrule
        shot-Free~\cite{RavichandranBS19} & ResNet12 & Train-DA  &59.04$\pm$\footnotesize0.43 & 77.64$\pm$\footnotesize0.39 &-&-&69.20$\pm$\footnotesize0.40 &84.70$\pm$\footnotesize0.40 &66.87$\pm$\footnotesize0.43 &82.64$\pm$\footnotesize0.39 \\
        MetaOpt~\cite{LiEDZW19} & ResNet12 & Train-DA & 62.64$\pm$\footnotesize0.61 & 78.63$\pm$\footnotesize0.46 & -&-&72.00$\pm$\footnotesize0.70 &84.20$\pm$\footnotesize0.50 &65.81$\pm$\footnotesize0.74 &82.64$\pm$\footnotesize0.39 \\
        Distill~\cite{TianWKTI20} & ResNet12 & Train\&test-DA\&KD &64.82$\pm$\footnotesize0.60 &82.14$\pm$\footnotesize0.43 &- &- &- &- &\textbf{71.52$\pm$\footnotesize0.69} &86.03$\pm$\footnotesize0.49  \\
        \textbf{HTS (ours)}  & ResNet12 & Train\&test-SSL &\textbf{64.95$\pm$\footnotesize0.18}  &\textbf{83.89$\pm$\footnotesize0.15} 
        &\textbf{72.88$\pm$\footnotesize0.22}  
        &\textbf{85.32$\pm$\footnotesize0.13} 
        &\textbf{73.95$\pm$\footnotesize0.22} 
        &\textbf{85.36$\pm$\footnotesize0.14} 
        &68.38$\pm$\footnotesize0.23 
        &\textbf{86.34$\pm$\footnotesize0.18} \\
        \bottomrule
      \end{tabular}}
  \end{center}
  \vspace{-8pt}
  \caption{Accuracy (\%) with \textbf{rotation3}. Best results are displayed in boldface. Train/test-SSL/-DA mean using data augmentation and self-superivised learning during training/testing. KD means knowledge distillation and \& means ``and'' operation.}
  \vspace{-28pt}
\label{Table:SOTA}
\end{table}

\vspace{-8pt}
\subsection{RQ2. Performance of Pretext Tasks in HTS}
\label{subsec:RQ2}
To answer RQ2, we conduct experiments on both single domain and cross domain with comparison to few-shot learning methods using the benchmark datasets.

\noindent \textbf{Performance on single domain.}
\cref{Table:SOTA} reports the average classification accuracy. 
For fairness, we use the same backbone network to compare to state-of-the-art meta-learning based FSL methods. 
From~\cref{Table:SOTA}, we have the following findings:
(1) HTS improves the performance of the ProtoNet in most settings to achieve a new state of the arts.
This is because that our framework has the advantage of modeling relationships among these images and adaptively learning augmented features via a gated selection aggregating component. 
This observation demonstrates our motivation and the effectiveness of our framework.
(2) One observation worth highlighting is that HTS not only outperforms traditional meta-learning based methods, but also is superior to methods using pretext tasks under DA or SSL settings.
(3) Compared to~\cite{TianWKTI20}, the results further show that augmenting the query set can bring more benefits during the testing phase.

\noindent \textbf{Performance on cross domain.}
In~\cref{Table:cross-domain}, we show the testing results on testing domains using the model trained on training domains.
The setting is challenging due to the domain gap between training and testing datasets. The results clearly show that 
(1) HTS has significant performance in all cross-domain settings and obtains consistent improvements.
(2) When CUB-200-2011 is selected as the testing domain, the transferred performance of different training domains has a large difference. It is caused by the degrees of domain gap.
From the last two rows, we find that all methods from \textit{tiered}ImageNet to \textit{mini}ImageNet have a large improvement compared with CUB-200-2011, because the two datasets come from the same database (ILSVRC-12), leading to a small domain gap.

\noindent \textbf{Performance with other meta-learning based methods.}
To further verify the effectiveness of HTS, we embed it into four meta-learning based methods: ProtoNet~\cite{SnellSZ17}, RelationNet~\cite{Sung2018}, MatchingNet~\cite{vinyals2016} and GNN~\cite{SatorrasE18}.
\cref{Table:adapting} reports the accuracy and we have the following findings:
(1) our HTS is flexible and can be combined with any meta-learning method, making the performance of these methods significantly improved in all datasets.
(2) In terms of our method, the HTS\_SSL performs better than  HTS\_DA in most cases, indicating that a single label space is not best for learning the knowledge carried by these pretext tasks.

\begin{figure}[tbp]
  \vspace{-2pt}
  \centering
  \subfigure[ProtoNet]{
    \includegraphics[width=2cm,height=1.5cm]{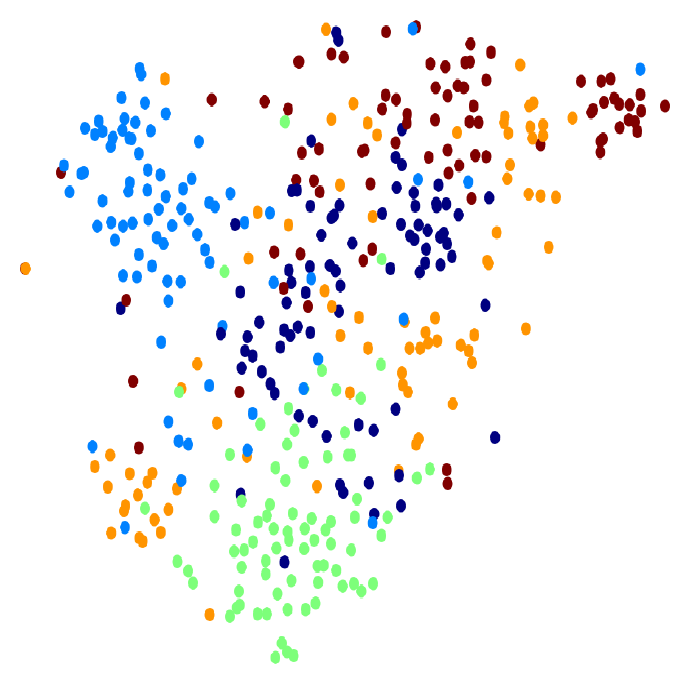}
    \label{fig:t-sne-a}
  }
  \subfigure[Backbone]{
    \includegraphics[width=2cm,height=1.5cm]{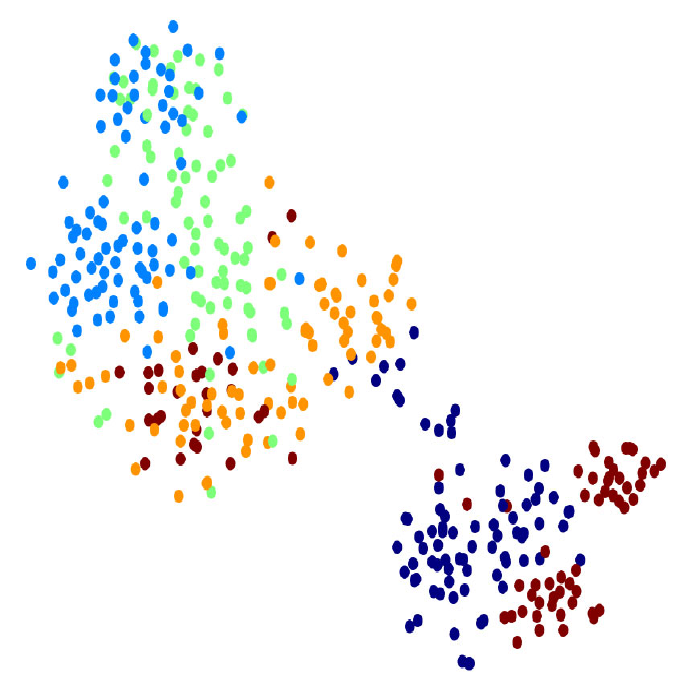}
    \label{fig:t-sne-b}
  }
  \subfigure[TreeLSTM]{
    \includegraphics[width=2cm,height=1.5cm]{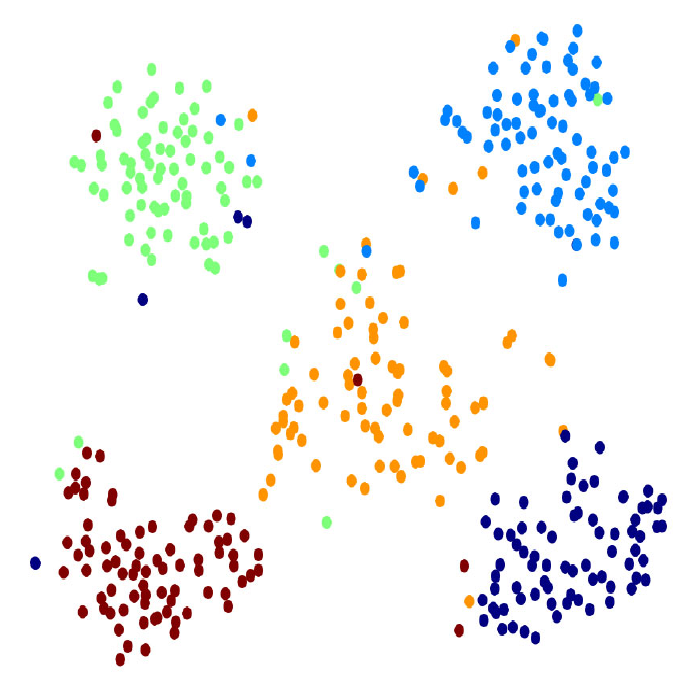}
    \label{fig:t-sne-c}
  }
  \vspace{-12pt}
  \caption{(a)-(c) represent the t-SNE of the five class features. (b) and (c) are our method.} 
  \label{fig:t-sne}
  \vspace{-18pt}
\end{figure}

\noindent \textbf{T-SNE visualization.} 
For the qualitative analysis, we also apply t-SNE~\cite{LaurensVisualizing2008} to visualize the embedding distribution obtained before and after being equipped with HTS in ProtoNet.
As shown in \cref{fig:t-sne}, (b) and (c) represent the features obtained without and with using the TreeLSTM aggregator.
The results show that (1) our HTS can learn more compact and separate clusters indicating that the learned representations are more discriminative.
(2) Considering our method, (c) is better than (b), which once again proves the effectiveness of the aggregator.

\begin{table}[tbp]
	\newcommand{\tabincell}[2]{\begin{tabular}{@{}#1@{}}#2\end{tabular}}
	\centering
	\resizebox{1.0\linewidth}{!}{
	  \begin{tabular}{cccccccccc}
		\toprule
		\multicolumn{2}{c}{\multirow{2}[4]{*}{\tabincell{c}{Methods \\ under 5-way}}} & \multicolumn{2}{c}{\textbf{\emph{mini}ImageNet}} &  \multicolumn{2}{c}{\textbf{CUB-200-2011}} & \multicolumn{2}{c}{\textbf{CIFAR-FS}} & \multicolumn{2}{c}{\textbf{\emph{tiered}ImageNet}} \\
		\cmidrule{3-10} &   & 1-shot & 5-shot & 1-shot & 5-shot & 1-shot & 5-shot & 1-shot & 5-shot  \\
		\midrule
		\multirow{3}[2]{*}{\tabincell{c}{ProtoNet \\ \cite{SnellSZ17}}}
		& & 44.42 & 64.24 & 51.31 & 70.77 & 51.90 & 69.95 & 49.35 & 67.28  \\
		& +HTS\_DA  & 57.39  & 74.25 & 66.88  & 77.37 & \textbf{65.01} \footnotesize\color{red}{(+13.11)} & 75.23 & 52.92  & 70.18\\
		& +HTS\_SSL  & \textbf{58.96} \footnotesize\color{red}{(+14.54)}     
		& \textbf{75.17} \footnotesize\color{red}{(+10.93)}
    & \textbf{67.32} \footnotesize\color{red}{(+16.01)}    
		& \textbf{78.09} \footnotesize\color{red}{(+7.32)}  & 64.71  
		&\textbf{76.45}\footnotesize\color{red}{(+6.50)}  &\textbf{53.20} \footnotesize\color{red}{(+3.85)}  &\textbf{72.38} \footnotesize\color{red}{(+5.10)}   \\
		\midrule
		\multirow{3}[2]{*}{\tabincell{c}{RelationNet \\ \cite{Sung2018}}} 
		&  & 49.31 &66.30 & 62.45 & 76.11 & 55.00 & 69.30 & 54.48 & 71.32  \\
		& +HTS\_DA  & 52.78  
    &73.22 &65.63  
    &\textbf{80.67} \footnotesize\color{red}{(+4.56)}  
		&57.68 &73.09  &55.44 &\textbf{72.88} \footnotesize\color{red}{(+1.56)}     \\
		& +HTS\_SSL   & \textbf{53.24} \footnotesize\color{red}{(+3.93)}    
    & \textbf{73.57} \footnotesize\color{red}{(+7.27)}   
    &\textbf{67.38} \footnotesize\color{red}{(+4.93)}    
		&79.00
		&\textbf{58.60} \footnotesize\color{red}{(+3.60)}   
		&\textbf{73.15} \footnotesize\color{red}{(+3.85)}   
		&\textbf{56.09} \footnotesize\color{red}{(+0.61)}  &71.78   \\
		\midrule
		\multirow{3}[2]{*}{\tabincell{c}{MatchingNet \\ \cite{vinyals2016}}} &  & 48.14 & 63.48 & 61.16 & 72.86 & 53.00  & 60.23  &54.02 & 70.11  \\
		& +HTS\_DA  & \textbf{53.64} \footnotesize\color{red}{(+5.50)}
    &63.67 &63.29   &73.17 &\textbf{55.57} \footnotesize\color{red}{(+2.57)} &62.58  &56.39  &72.01   \\
		& +HTS\_SSL & 52.29
		&\textbf{64.36} \footnotesize\color{red}{(+0.88)}
    &\textbf{63.54} \footnotesize\color{red}{(+4.93)}   
		&\textbf{74.76} \footnotesize\color{red}{(+1.90)}   
    &55.18 
		&\textbf{63.87} \footnotesize\color{red}{(+3.64)}
    &\textbf{57.16} \footnotesize\color{red}{(+3.14)} 
		&\textbf{72.60} \footnotesize\color{red}{(+2.49)}   \\
		\midrule
		\multirow{3}[2]{*}{\tabincell{c}{GNN \\ \cite{SatorrasE18}}} &  & 49.02 
		& 63.50 & 51.83 & 63.69 & 45.59 & 65.62 & 43.56 & 55.31  \\
		& +HTS\_DA   & 59.06   &72.80  &61.69   &73.46 &52.35 &71.66  &55.32  &69.48  \\
		& +HTS\_SSL  &\textbf{60.52} \footnotesize\color{red}{(+11.50)}
		&\textbf{74.63} \footnotesize\color{red}{(+11.13)}  
    &\textbf{62.85} \footnotesize\color{red}{(+11.02)}
		&\textbf{77.58} \footnotesize\color{red}{(+13.89)} 
    &\textbf{58.31} \footnotesize\color{red}{(+12.72)}    
		&\textbf{73.24} \footnotesize\color{red}{(+7.62)}    
    &\textbf{55.73} \footnotesize\color{red}{(+12.17)}
		&\textbf{70.42}\footnotesize\color{red}{(+15.11)}    \\
		\bottomrule
	\end{tabular}}
	\vspace{-2pt}
	\caption{Accuracy (\%) by incorporating HTS (two-level trees) into each method with rotation3 and the Conv4-64. Best results are displayed in \textbf{boldface} and performance improvement in red text. \cref{asec:models} shows the formulation of these methods.}
\label{Table:adapting}
\vspace{-22pt}
\end{table}

\subsection{RQ3. Adaptive Selection Aggregation}
\label{subsec:RQ3}

\begin{figure}[tbp]
  \vspace{-5pt}
  \centering
  \includegraphics[width=0.225\textwidth]{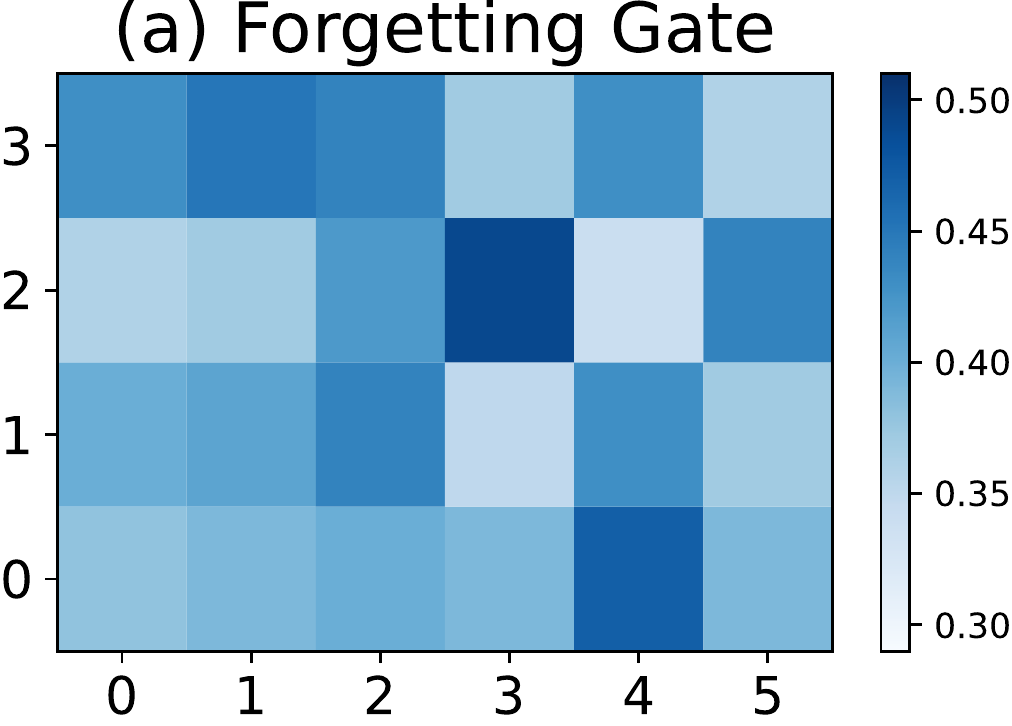}
  \label{fig:RQ3-a}
  \includegraphics[width=0.225\textwidth]{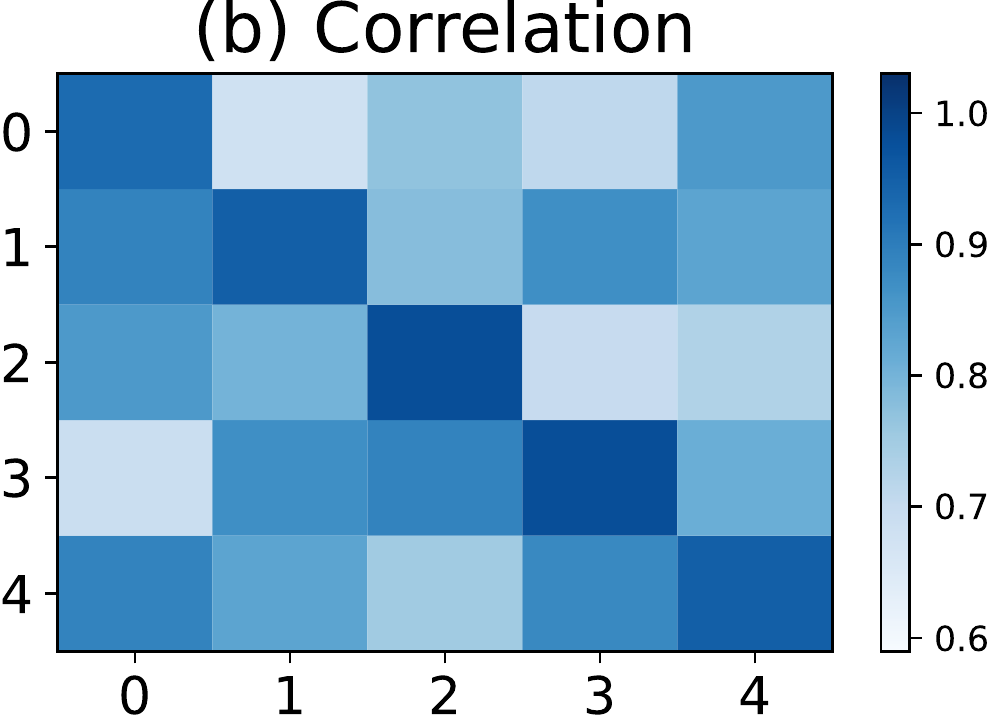}
  \label{fig:RQ3-b}
  \includegraphics[width=0.25\textwidth]{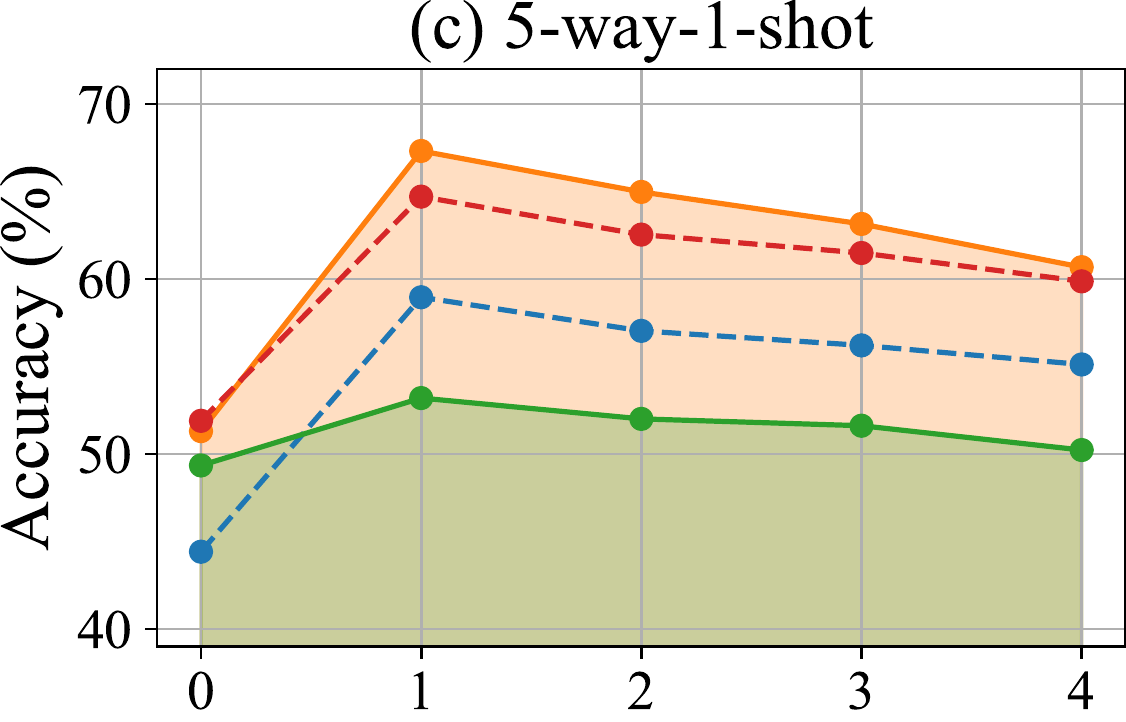}
  \label{fig:dnl-a}
  \includegraphics[width=0.25\textwidth]{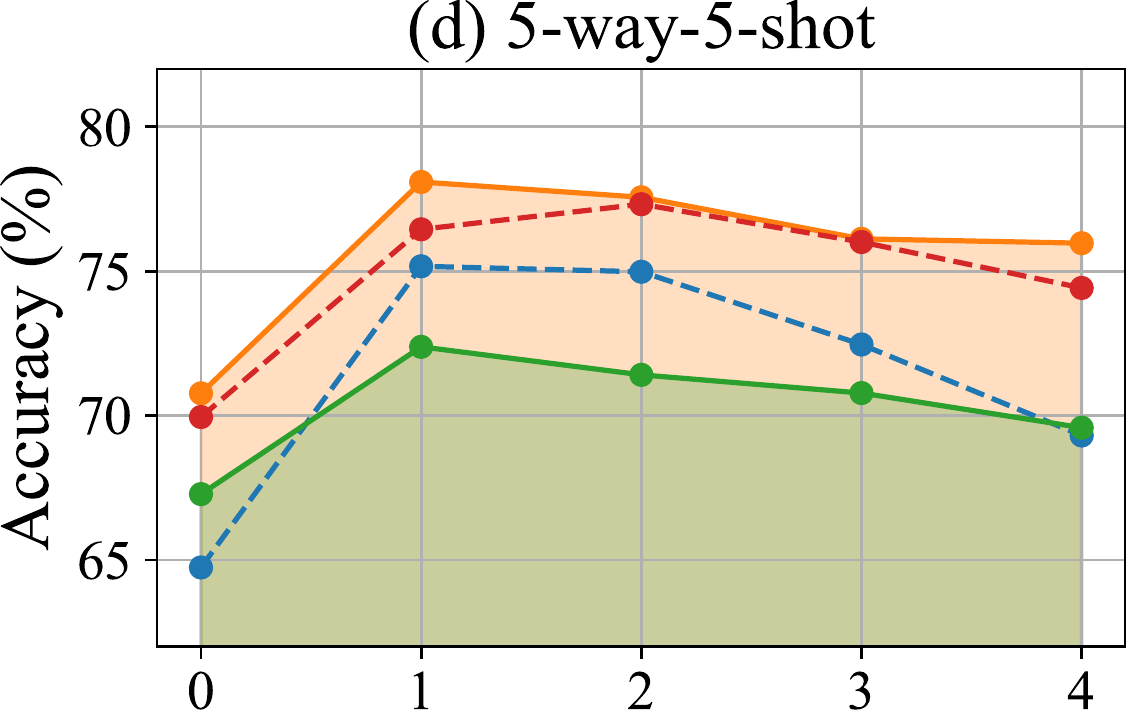} \\
  \includegraphics[width=0.8\textwidth]{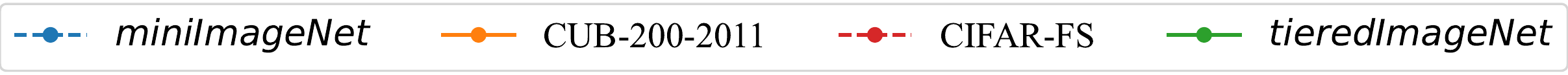}
  \vspace{-10pt}
  \caption{(a) is the forgetting gate for different child nodes (vertical axis) with the same raw image (horizontal axis). 
  (b) is the correlations of different augmented images (vertical axis is rotation1,  horizontal axis is color\_perm1), where the diagonal (or off-diagonal) line is based on the same raw (or different raw) images.
  (c)-(d) are classification accuracy (\%) for different numbers of levels under ProtoNet with HTS.} 
  \label{fig:dnl}
  \vspace{-18pt}
\end{figure}

One of the most important properties of our framework is that the learned augmented features are different for different pretext tasks or child nodes.
Thus, in this subsection, we investigate if the proposed framework can adaptively learn different forgetting gates for different child nodes, which aims to answer RQ3.
We visualize the average forgetting value (\textit{i.e.}, $f_{m}$ in~\cref{Eq:treelstm}) for each child node with the same raw image randomly sampled from CUB-200-2011.
In~\cref{fig:dnl} (a), we show the results of six raw images as we have similar observations on other images and datasets.
The darkness of a step represents the probability that the step is selected for aggregation.
We can find that (1) different types of child nodes exhibit different forgetting values based on the same raw image.
(2) Different raw images focus on different child node features.
These results further explain that the contributions of different pretext tasks to the raw images are not equal.
In~\cref{fig:dnl} (b), we also provide the correlation, \textit{i.e.}, the cosine similarity of learned features of different types of augmentation based on the same raw image. 
We clearly observe the correlations between augmentation with the same raw image are large while the correlation between different raw images is small, which meets our expectation that multiple sets of augmentation generated based on the same raw image generally have similar relationships.

\vspace{-3pt}
\subsection{RQ4. Ablation Study}
\label{subsec:RQ4}
\vspace{-3pt}
To answer RQ4, we conduct experiments to show the performance under different pretext tasks, child nodes and levels.  
More experiments see~\cref{asec:RQ4}. 

\begin{figure}[tbp]
  \centering
  \includegraphics[width=0.235\textwidth]{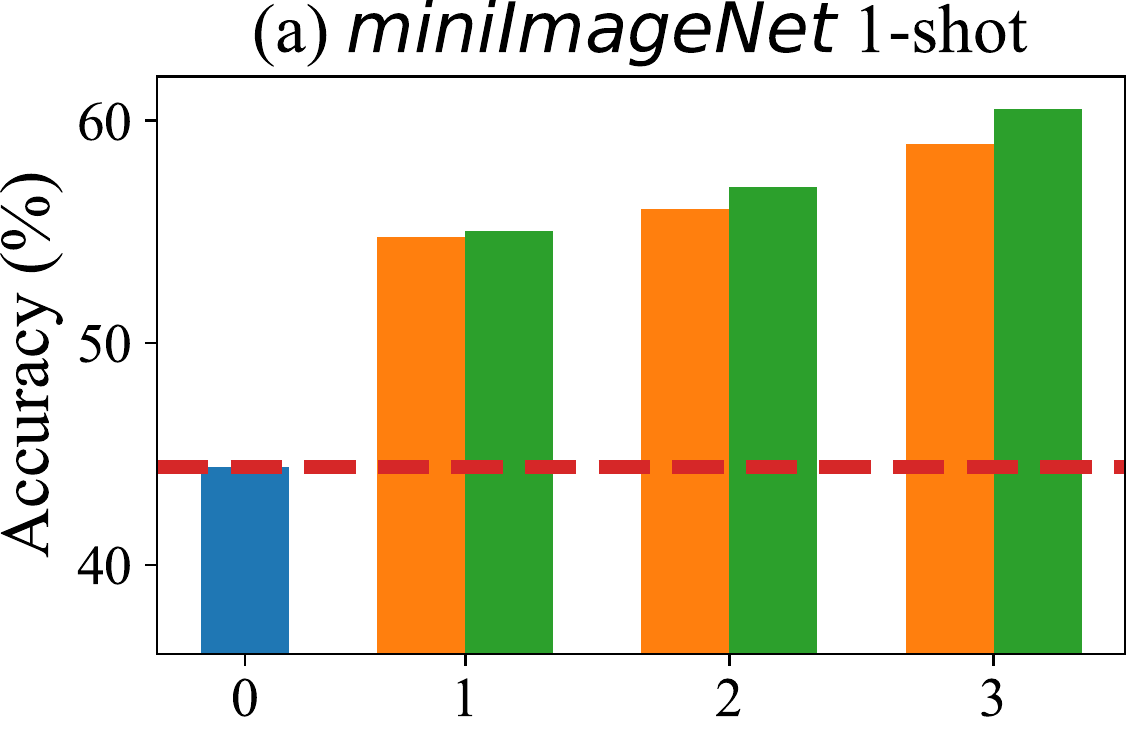}
  \label{fig:dtt-a}
  \includegraphics[width=0.235\textwidth]{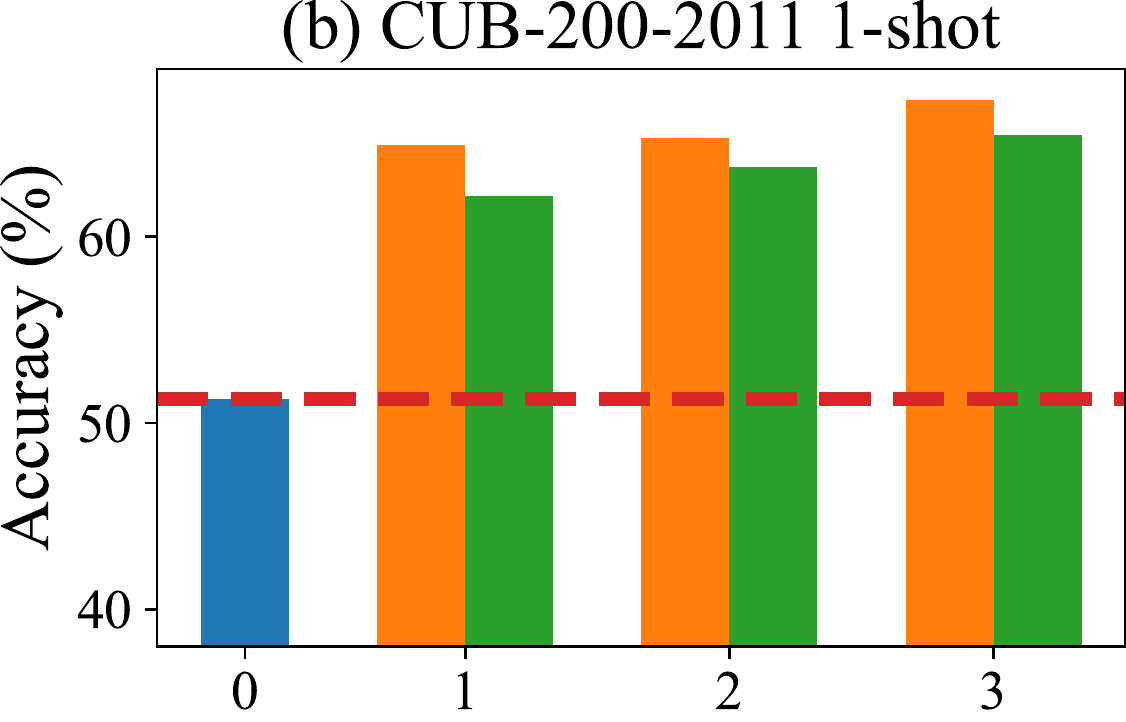}
  \label{fig:dtt-b}
  \includegraphics[width=0.235\textwidth]{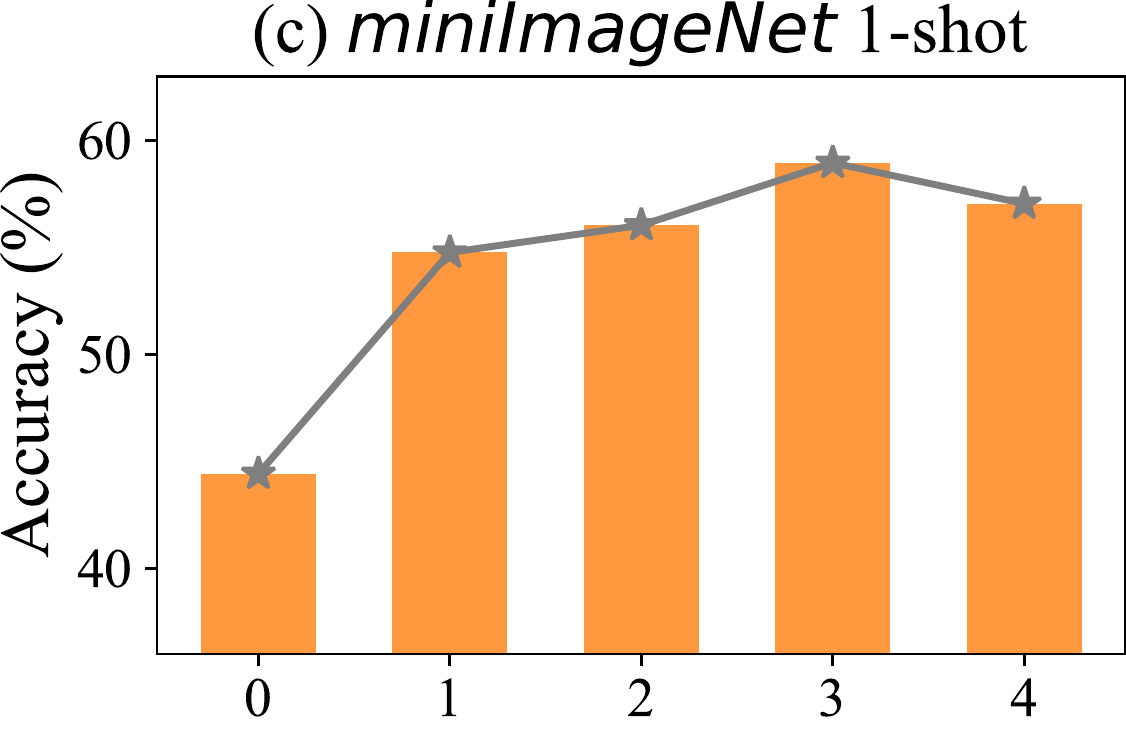}
  \label{fig:dcn-a}
  \includegraphics[width=0.235\textwidth]{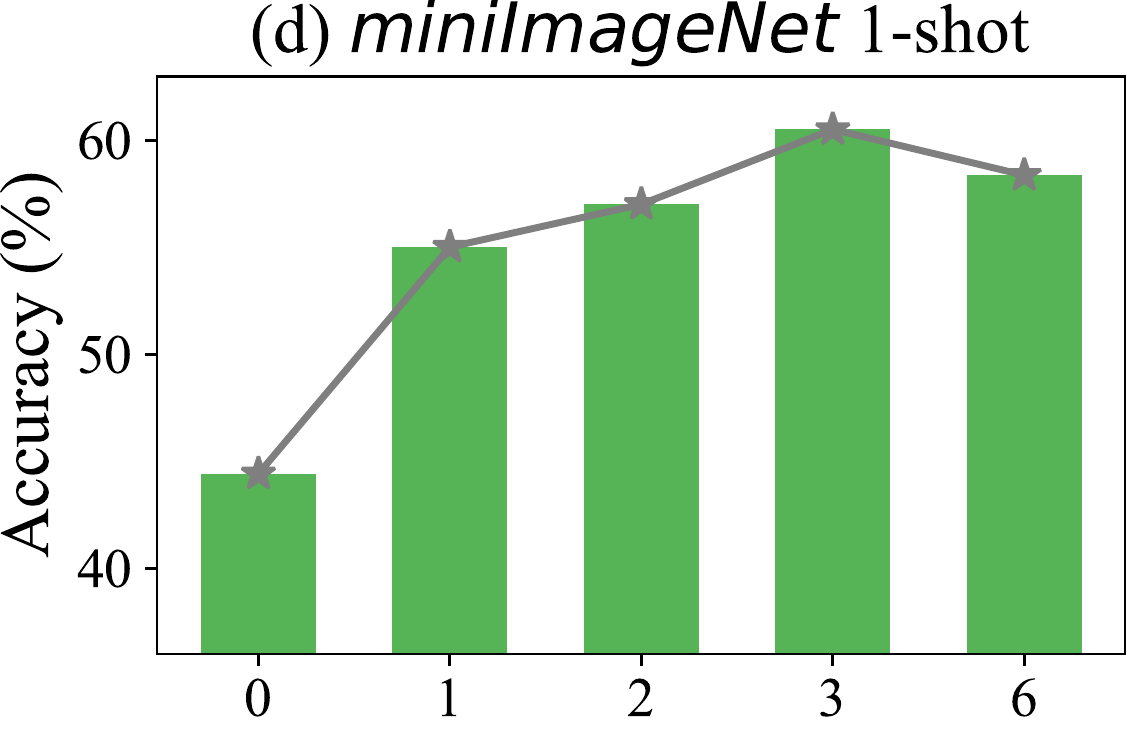}
  \label{fig:dcn-b} \\
  \includegraphics[width=0.45\textwidth]{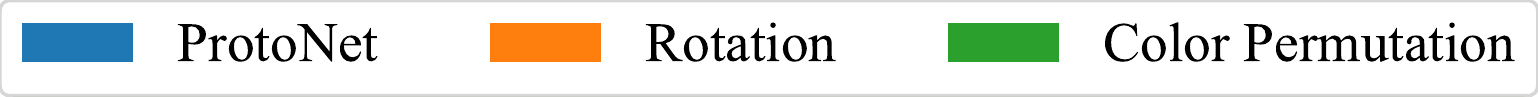}
  \vspace{-13pt}
  \caption{(a) and (b) indicate different pretext tasks with the same number of child nodes. (c) and (d) indicate different numbers of child nodes with same pretext task.} 
  \label{fig:dtt}
\vspace{-16pt}
\end{figure}

\noindent \textbf{The effect of different pretext tasks.}
To show the performance of different pretext tasks, we construct experiments based on the two-level trees (\textit{i.e.} only using a pretext task) for each specific pretext task with the same number of child nodes on the \textit{mini}ImageNet and CUB-200-2011 under the 5-way 1-shot setting.
\cref{fig:dtt} (a) and (b) show the results.
Along the horizontal axis, the number 0 indicates the baseline (\textit{i.e.}, ProtoNet) only using raw (FSL) images. 
The rotation1/2/3 and color\_perm1/2/3 are used for the numbers 1/2/3, respectively.
We follow these findings:
(1) all pretext tasks significantly improve the classification performance and outperform the baseline.
These results indicate that our HTS does not need to manually define a suitable pretext task, and it can adaptively learn useful information. 
(2) The color permutation task brings a higher boost on \textit{mini}ImageNet and the rotation task has a good performance on CUB-200-2011.
It is because for fine-grained image classification, rotation angles can carry affluent representations, but for a large of images, the color gain is higher.

\noindent \textbf{The effect of different numbers of child nodes.}
We verify the performance of the same pretext task with different numbers of child nodes on the \textit{mini}ImageNet under the 5-way 1-shot setting.
\cref{fig:dtt} (c) and (d) show the results.
Along the horizontal axis, these integers are introduced in~\cref{subsec:setup}. 
From~\cref{fig:dtt} (c)-(d), we have the following findings: (1) our proposed HTS method with the two-level trees outperforms the ProtoNet in terms of all settings, which demonstrates the effectiveness of our method to learn pretext tasks.
(2) The performance with different numbers of child nodes is different.
The reason is that the increase of child nodes is accompanied by large model complexity, leading to overfitting and affecting the generalization ability of the FSL model. 

\noindent \textbf{The effect of different numbers of tree levels.}
We further verify the performance of HTS with different numbers of levels on the four datasets under the 5-way 1-shot and 5-shot settings.
\cref{fig:dnl} (c) and (d) show the results.
Along the horizontal axis, the integer 1 indicates only using FSL images, 2 using [rotation3], 3 using [rotation3; color\_perm2], 4 using [rotation3; color\_perm2; color\_perm3] and 5 using [rotation3; color\_perm2; color\_perm3; rotation2].
We find that our HTS method with different tree levels outperforms the original baseline (ProtoNet) using only raw (FSL) images.
The different number of tree levels bring model complexity and over-fitting problems, resulting in performance differences among different levels.
To save computational time and memory, two-level trees are used, which achieves the new state-of-the-art performance.

\section{Conclusion}
\label{sec:con}
\vspace{-5pt}
In this paper, we study the problem of learning richer and transferable representations for few-shot image classification via pretext tasks. We propose a plug-in \textit{hierarchical tree structure-aware (HTS)} method that constructs tree structures to address this problem. Specifically, we propose the hierarchical tree constructing component, which uses the edge information to model the relationships of different augmented and raw images, and uses the level to explore the knowledge among different pretext tasks. On the constructed tree structures, we also introduce the gated selection aggregating component, which uses the forgetting gate to adaptively select and aggregate the augmented features and improve the performance of FSL. Extensive experimental results demonstrate that our HTS can achieve a new state-of-the-art performance on four benchmark datasets.

\
\newline
\noindent \textbf{Acknowledgments:} This work was supported by the National Science and Technology Innovation 2030 - Major Project (Grant No. 2022ZD0208800), and NSFC General Program (Grant No. 62176215).
We thank Dr. Zhitao Wang for helpful feedback and discussions.

\clearpage
\section*{Appendix}

\appendix
\renewcommand\thesection{A.\arabic{section}}
\renewcommand\thefigure{A.\arabic{figure}}   
\renewcommand\thetable{A.\arabic{table}}  
\renewcommand\theequation{A.\arabic{equation}}

\noindent The supplementary material is organized as:

1. The full algorithm for few-shot image classification with HTS is shown in~\cref{asec:alg}. 

2. The formulations of meta-learning based FSL methods are shown in~\cref{asec:models}. 

3. Additional experimental setups are shown in~\cref{asec:set}.

4. Additional experimental results for RQ2 (main paper) are shown in~\cref{asec:RQ2}.

5. Additional experimental results for RQ4 (main paper) are shown in~\cref{asec:RQ4}.

\section{Full HTS Algorithm} 
\label{asec:alg}
For easy reproduction, we present the full algorithm for few-shot image classification with HTS in~\cref{Algorithm:HTS}.
Once trained, with the learned model parameters, we can perform the meta-testing phase over the test episodes with~\cref{Eq:test}.

\section{Meta-Learning based FSL Models}
\label{asec:models}
In the main paper, we have introduced the ProtoNet~\cite{SnellSZ17} as the FSL model of the HTS in~\cref{subsec:class}.
Due to the flexibility of the HTS (\textit{i.e.}, it can be combined with any meta-learning based FSL model), we also show the experimental results obtained by using other popular meta-learning based models in~\cref{subsec:RQ2}. 
Next, we detail these methods how work with the proposed HTS method.

\begin{algorithm}[tbp]
	\centering
	\footnotesize
	\caption{Hierarchical Tree Structure-Aware Method}
	\begin{algorithmic}[1]
		\REQUIRE The training set $\mathcal{D}_{b}$, the pretext-task operators $\mathcal{G}$ \\
    \qquad \ \, The loss weight hyperparameter $\{\beta_{j}|j=1,\cdots, J\}$
		\STATE Randomly initialize all learnable parameters $\{\phi, \theta^{r}|r=0,\cdots, J\}$
		\FOR{iteration=1,$\cdots$, MaxIteration}
		\STATE Randomly sample episode $\mathcal{T}_{e}$ from $\mathcal{D}_{b}$
    \STATE Generate the set of augmented episodes $\mathcal{T}_{agg}$ from  $\mathcal{T}_{e}$ using  $\mathcal{G}$
    \STATE Extract image features of $\mathcal{T}_{e}$ and $\mathcal{T}_{agg}$ using shared encoder $E_{\phi}$
    \STATE // \textbf{Hierarchical tree constructing component}
    \STATE The form of the tree \{$E_{\phi}(x_{i})\overset{g_{1}}{\to}E_{\phi}(x_{i}^{1})\overset{g_{2}}{\to}\cdots\overset{g_{j}}{\to}E_{\phi}(x_{i}^{j})\cdots\overset{g_{J}}{\to}E_{\phi}(x_{i}^{J})$\} is used to construct tree structure
    \STATE  // \textbf{Gated selection aggregating component}
    \STATE The learning process of the tree \{$h_{i}^{0}\overset{agg}{\longleftarrow}h_{i}^{1}\overset{agg}{\longleftarrow}\cdots\overset{agg}{\longleftarrow}h_{i}^{r}\cdots\overset{agg}{\longleftarrow}E_{\phi}(x_{i}^{J})$\} is used to hierarchically aggregate node information
    \STATE // \textbf{Meta-training phase}
		\IF {Data augmentation setting}
		\STATE Update the FSL model using~\cref{Eq:hts-da};
		\ELSIF {Self-supervised learning setting} 
		\STATE Update the FSL model using~\cref{Eq:hts-ssl};
		\ENDIF 
		\ENDFOR
	\end{algorithmic}
\label{Algorithm:HTS}
\end{algorithm}

\noindent \textbf{Graph Neural Network (GNN).}
GNN~\cite{SatorrasE18,zhao2022graph,zhao2021multimodal,zhao2021multi} uses the posterior inference over a graphical model determined by the inputs and labels to formally propagate information. 
It constructs a fully-connected graph $G_{\mathcal{T}_{e}}=(V,E)$, where nodes $v_{n}\in V$ correspond to the images present in each episode $\mathcal{T}_{e}=\{\mathcal{S}_{e}, \mathcal{Q}_{e}\}$. $e_{n,n^{'}}\in E$ represents the similarity of image $x_{n}$ and $x_{n^{'}}$ and uses a parametric model (\textit{i.e.} MLP) to learn.
GNN contains a feature encoder $E_{\phi}$ with learnable parameters $\phi$ (\textit{e.g.}, CNN) to extract the image features. 
Then these features are used to calculate the similarities between images using an MLP network.
Over the weighted graph $G$, a GNN is used to propagate and update the node information.
Finally, updated feature representations of query images are input a classifier to predict the labels.
The training process of GNN is shown as:
\begin{equation} \label{Eq:gnn_1} 
	\begin{aligned}
	  &L_{FSL}(\mathcal{S}_{e}, \mathcal{Q}_{e}) = \frac{1}{|\mathcal{Q}_{e}|}\sum\nolimits_{(x_{i},y_{i}\in \mathcal{Q}_{e})} -log \ p_{y_{i}}, \\
	  &p_{y_{i}} = GNN(MLP(E_{\phi}(\mathcal{S}_{e}), E_{\phi}(x_{i}))). \\
	\end{aligned}
\end{equation}

GNN adopts the message propagating process, which iteratively aggregates the neighborhood information (\textit{i.e.}, support labeled information). 
Formally, the aggregation and propagation process of the $k$-th layer in GNN is two steps: 
\begin{equation} \label{Eq:gnn_2} 
	\begin{aligned}
	  &m_{k,n} = AGGREGATE(\{a_{k-1,u}:u \in \mathcal{N}_{(n)}\}), \\
	  &a_{k,n} = UPDATE(a_{k-1},m_{k,n},a_{0,n}), \\ 
	\end{aligned}
\end{equation}
where $\mathcal{N}_{n}$ is the set of neighbors of node $v_{n}$, $a_{0,n}$ is initial features and AGGREGATE is a permutation invariant function.
After K message-passing layers, the final node embeddings $A_{K}$ are used to infer the query image label. 
GNN combined with the proposed HTS method only need to replace $L_{FSL}$ (\cref{Eq:hts-da} - \cref{Eq:test} in the main paper) with \cref{Eq:gnn_1}.

\noindent \textbf{Matching Network (MatchingNet).}
MatchingNet~\cite{vinyals2016} contains a feature encoder $E_{\phi}$ and a attention or memory network.
In each episode $\mathcal{T}_{e} =\{\mathcal{S}_{e}, \mathcal{Q}_{e}\}$, MatchingNet labels each query image as a consine distance-weighted linear combination of the support labels and uses a attention mechanism to calculate.
The training process of MatchingNet is shown as:
\begin{equation} \label{Eq:matching} 
	\begin{aligned}
	  &L_{FSL}(\mathcal{S}_{e}, \mathcal{Q}_{e}) = \frac{1}{|\mathcal{Q}_{e}|}\sum\nolimits_{(x_{i},y_{i}\in \mathcal{Q}_{e})} -log \ p_{y_{i}}, \\
	  &p_{y_{i}} = \sum\nolimits_{\hat{x}_{i}, \hat{y}_{i}\in \mathcal{S}_{e}}a(x_{i}, \hat{x}_{i})\cdot\mathbb{I}(\hat{y}_{i}=y_{i}),
	\end{aligned}
\end{equation}
where $a(\cdot,\cdot)$ represents an attention mechanism and it fully specifies the classifier. $\mathbb{I}$ represents the indicator function with its output being 1 if the input is true or 0 otherwise. Our proposed HTS method with MatchingNet only need to replace $L_{FSL}$ (\cref{Eq:hts-da} - \cref{Eq:test} in the main paper) with \cref{Eq:matching}.

\noindent \textbf{Relation Network (RelationNet).}
RelationNet~\cite{Sung2018} is comprised of a feature encoder $E_{\phi}$ as usual, and a relation module $R_{\omega}$ parameterized by $\omega$. 
They first embed each support and query using $E_{\phi}$ and create a prototype $\tilde{p}_{c}$ for each class $c$ by averaging its support embeddings, as shown in Eq.(1) in the main paper. 
Each prototype $\tilde{p}_{c}$ is concatenated with each embedded query and fed through the relation module which outputs a number in $[0, 1]$ representing the predicted probability that that query belongs to class $c$. 
The query loss is then defined as the mean square error of that prediction compared
to the (binary) ground truth. Both the encoder and the relation module are trained to minimize this loss.
\begin{equation} \label{Eq:relation} 
	\begin{aligned}
	  &L_{FSL}(\mathcal{S}_{e}, \mathcal{Q}_{e}) = \frac{1}{|\mathcal{Q}_{e}|}\sum\nolimits_{(x_{i},y_{i}\in \mathcal{Q}_{e})} -log \ p_{y_{i}}, \\
	  &p_{y_{i}} = \sum\nolimits_{c\in \mathcal{C}_{e}}(r_{c,i}-\mathbb{I}(y_{i} = y_{c}))^{2}, \\
	  &r_{c,i} = R_{\omega}(Ca(\tilde{p}_{c},E_{\phi}(x_{i}))),
	\end{aligned}
\end{equation}
where $Ca(\cdot,\cdot)$ is a concatenated operator. The proposed HTS with MatchingNet only need to replace $L_{FSL}$ (\cref{Eq:hts-da} - \cref{Eq:test} in the main paper) with~\cref{Eq:relation}.

\section{Additional Experimental Setup} 
\label{asec:set}
\noindent \textbf{Benchmark Datasets.}
Four benchmark few-shot learning datasets are used to evaluate our HTS: 
(1) \textbf{\emph{mini}ImageNet} contains 600 images per class over 100 classes. These classes are divided into 64/16/20 for train/val/test~\cite{vinyals2016}. 
(2) \textbf{\emph{tiered}ImageNet} is much larger compared with \emph{mini}ImageNet with 608 classes and each class about 1,300 samples. These classes were grouped into 34 higher-level concepts and then partitioned into 20/6/8 disjoint sets for train/val/test to achieve a larger domain difference between training and testing phases~\cite{RenTRSSTLZ18}.  
(3) \textbf{CUB-200-2011} is initially designed for fine-grained classification. It contains 200 classes and each class has around 60 samples. We divided these classes into 100/50/50 for train/val/test following the previous works~\cite{WahCUB_200_2011} . 
(4) \textbf{CIFAR-FS} is a subset of CIFAR-100 and has 600 images per class over 100 classes. The train/val/test classes are same to \emph{mini}ImageNet~\cite{BertinettoHTV19}. 
We also list the images number, classes number, images resolution and train/val/test splits following the criteria of previous works in~\cref{Table:datasets}. 

\begin{table}[htbp]
	\begin{center}
		\begin{tabular}{ccccc}
			\toprule
			Datasets & Images & Classes & Train/Val/Test & Resolution \\
			\textit{mini}ImageNet & 60000 & 100 & 64/16/20 & 84$\times$84 \\
			\textit{tiered}ImageNet & 779165 & 608 & 351/97/160 & 84$\times$84 \\
			CUB-200-2011 & 11788 & 200 & 100/50/50 & 84$\times$84  \\
			CIFAR-FS & 60000  & 100 & 64/16/20  & 32$\times$32 \\
			\bottomrule
		\end{tabular}
	\end{center}
	\caption{The details for four benchmark datasets.}
	\label{Table:datasets}
	\vspace{-12pt}
\end{table}

\noindent \textbf{Implementation Details.}
Following~\cite{Huang2021AGAM}, we apply Conv4 and ResNet12 as the convolution backbone networks to fairly compare the results.
The Conv4 consists of a stack of four Conv-BN-ReLU convolutional blocks and each convolutional block with 64 filters.
ResNet12 mainly has four blocks, which include one residual block. 
The last features of all backbone networks are processed by a global average pooling, then followed by a fully-connected layer with batch normalization~\cite{ioffe2015batch} to obtain a 64-dimensions instance embedding.
We use Adam optimizer with an initial learning rate of $10^{-3}$, and reduce the learning rate by 15K episodes of all datasets.
The weight decay is set to $5e^{-4}$.

\section{Additional Experiments for RQ2} 
\label{asec:RQ2}
In this section, we report additional experiments to further prove the effectiveness of our proposed HTS method. 

\begin{figure}[htbp]
	\centering
	\includegraphics[width=0.80\linewidth]{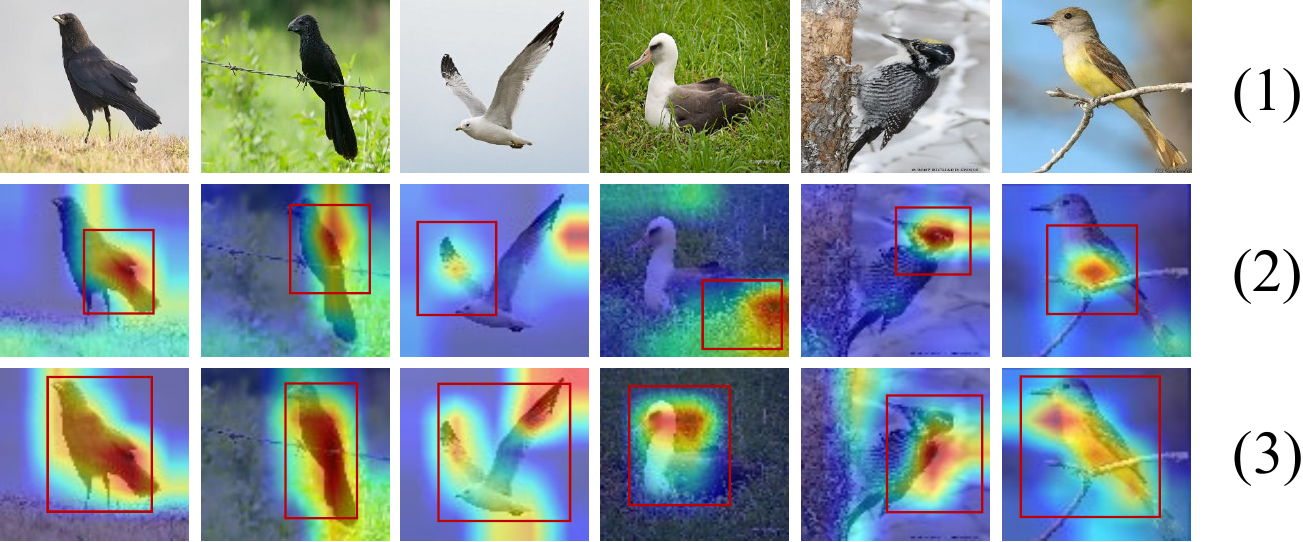}
	\caption{Grad-CAM visualizations of CUB-200-2011 dataset. Each column shows the result of same test image, and each row shows: (1) The results of test images. (2) The results of ProtoNet. (3) The results of our HTS. Best viewed in color and our proposed HTS method achieves better results for all test images.} 
	\label{fig:CAM} 
\end{figure}

\noindent \textbf{Visualization of CAM.} 
To evaluate whether a model exploits an important or actual object when making a prediction, \cref{fig:CAM} visualizes the gradient-weighted class activation mapping (Grad-CAM)~\cite{SelvarajuCDVPB17} from ProtoNet and HTS under a Conv4-64 feature encoder.
It is observed that using HTS makes the model pay more attention to the local object features than the ProtoNet.
It also further shows that the hierarchical aggregated representations can avoid semantic bias caused by pretext tasks.
Therefore, the proposed HTS helps the meta-learning based methods to use the correct visual features for prediction.

\begin{figure}[htbp]
	\centering
	\vspace{-10pt}
	\includegraphics[width=0.235\textwidth]{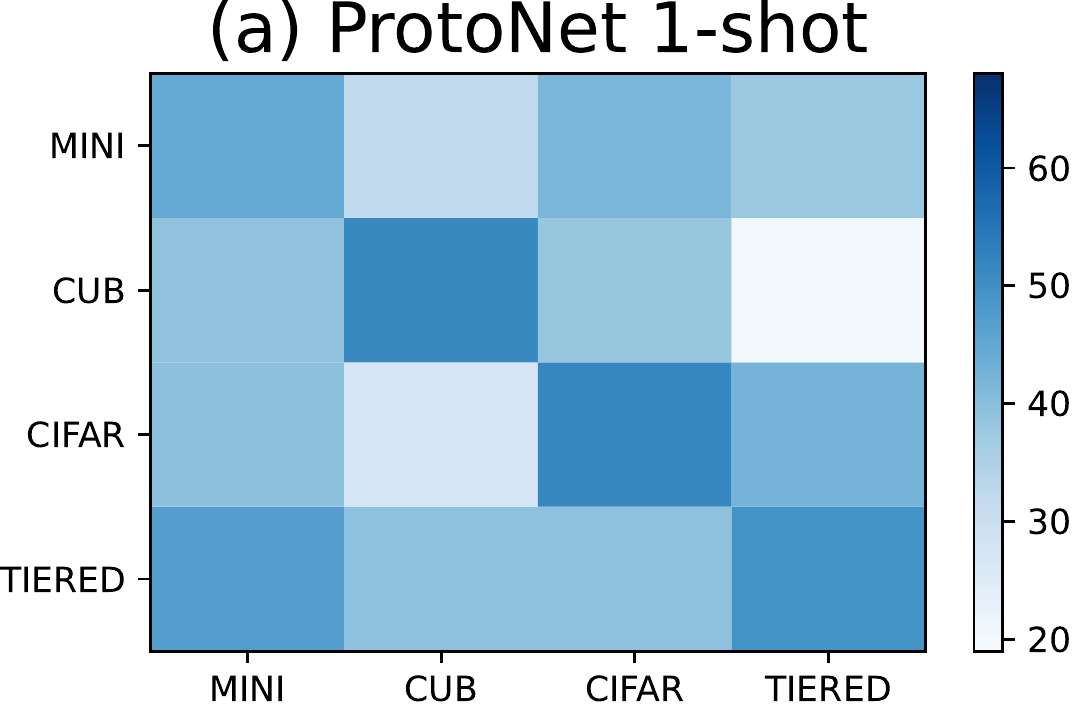}
    \includegraphics[width=0.235\textwidth]{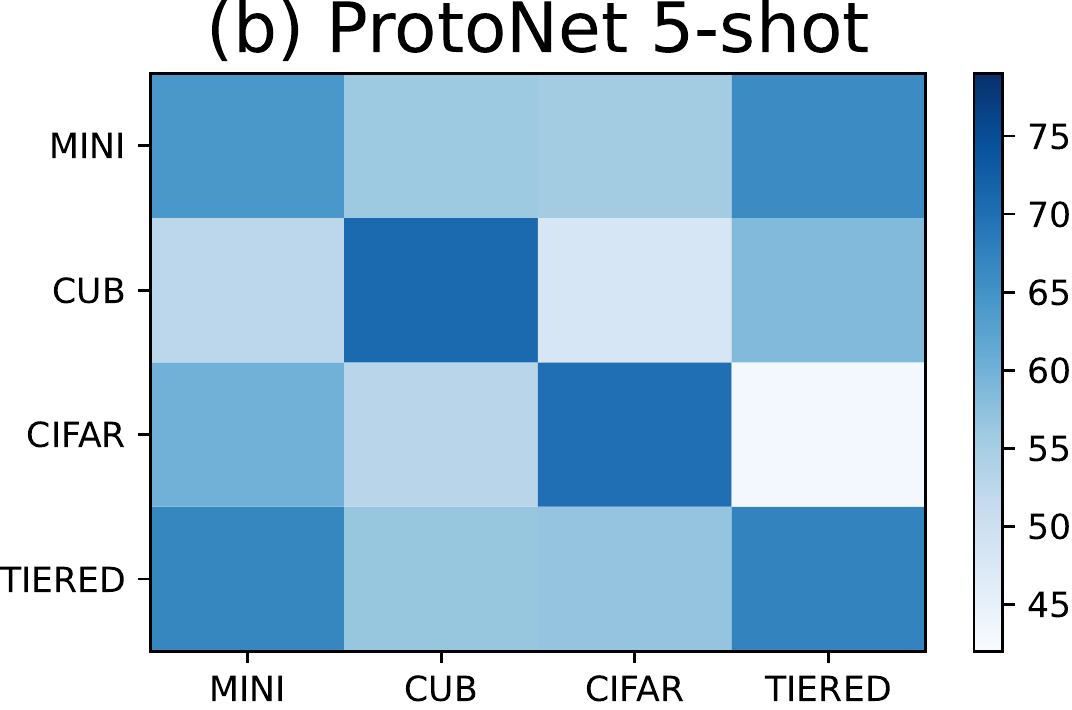}
	\includegraphics[width=0.235\textwidth]{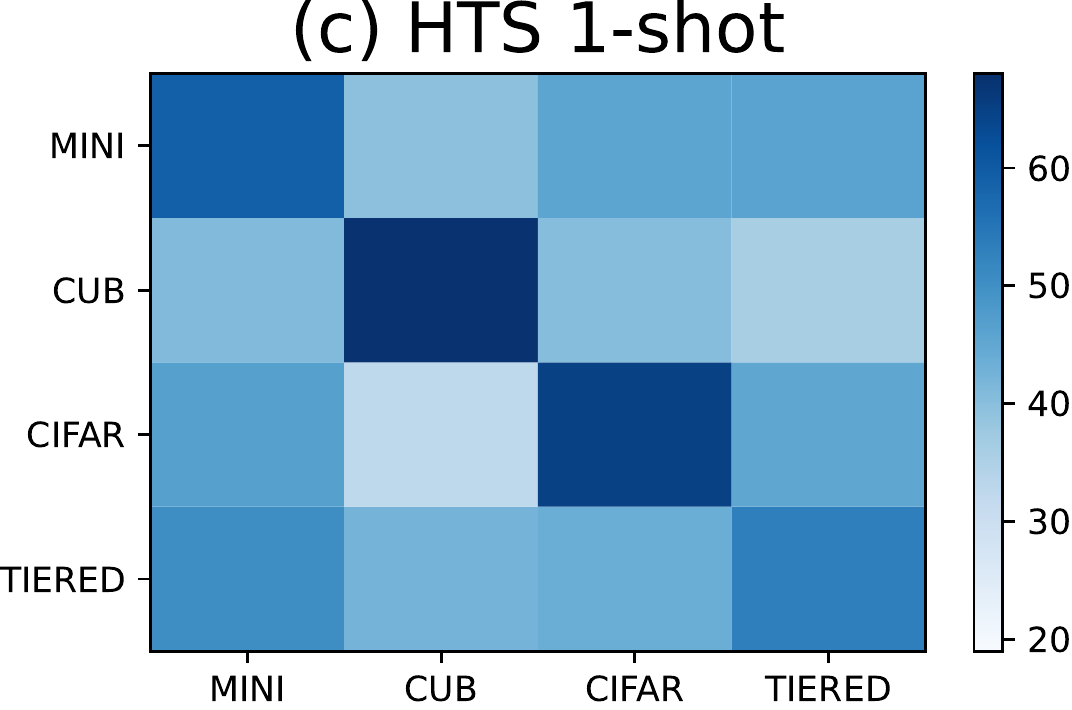}
	\includegraphics[width=0.235\textwidth]{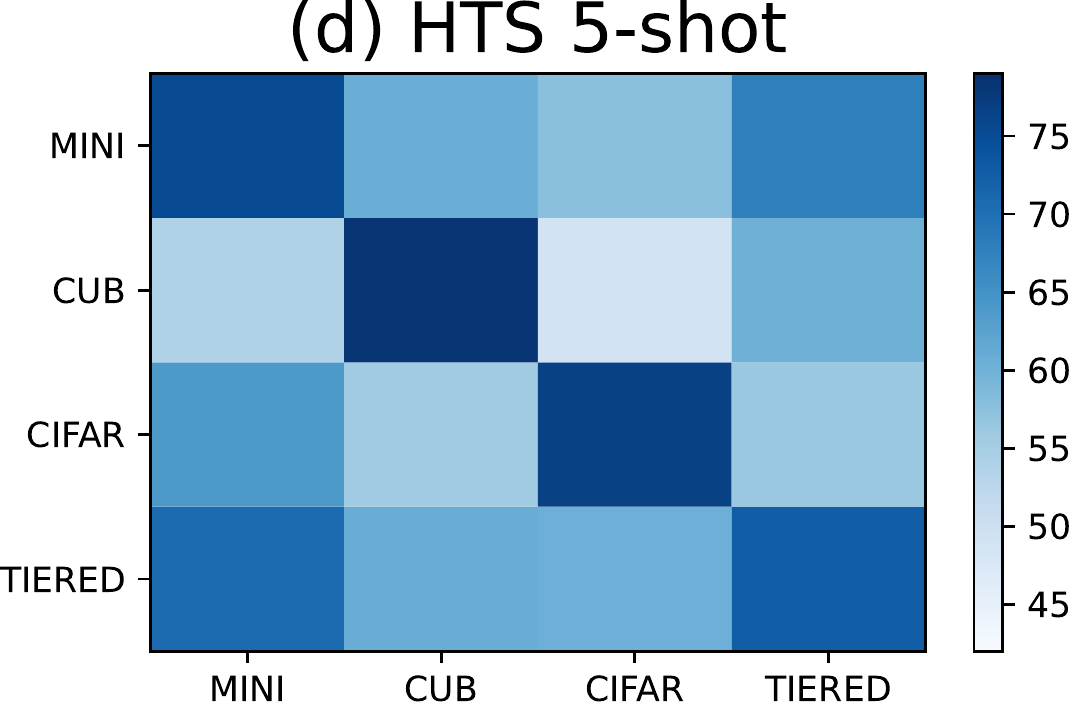}
	\vspace{-5pt}
	\caption{Cross-domain evaluation with \textbf{rotation3} under 5-way 1-shot and 5-shot settings. The horizontal axis represents training domains and the vertical axis is testing domains on miniImagenet (MINI), CUB-200-2011 (CUB), CIFAR-FS (CIFAR) and tieredImagenet (TIERED). To clearly show the performance, we use the same color bar with the same number of shots under baseline and our method.} 
	\label{fig:cross_domain}
\end{figure}

\noindent \textbf{Performance on cross domain.}
In~\cref{fig:cross_domain}, we give the cross-domain performance of four datasets as the training and testing domains respectively.
To clearly show the performance, we use the same color bar with the same number of shots based on baseline and our method.
From~\cref{fig:cross_domain}, our method outperforms the ProtoNet under all domain settings including singe domain (diagonal line) and cross domain (off-diagonal line).
This indicates that our HTS can improve the generalization ability of the FSL model with only a few labeled images.

\section{Additional Experimental for RQ4}
\label{asec:RQ4}

More experiment results are shown under different pretext tasks with the same child nodes to supplement RQ4.

\begin{figure}[htbp]
	\centering
	\vspace{-30pt}
	\includegraphics[width=0.235\textwidth]{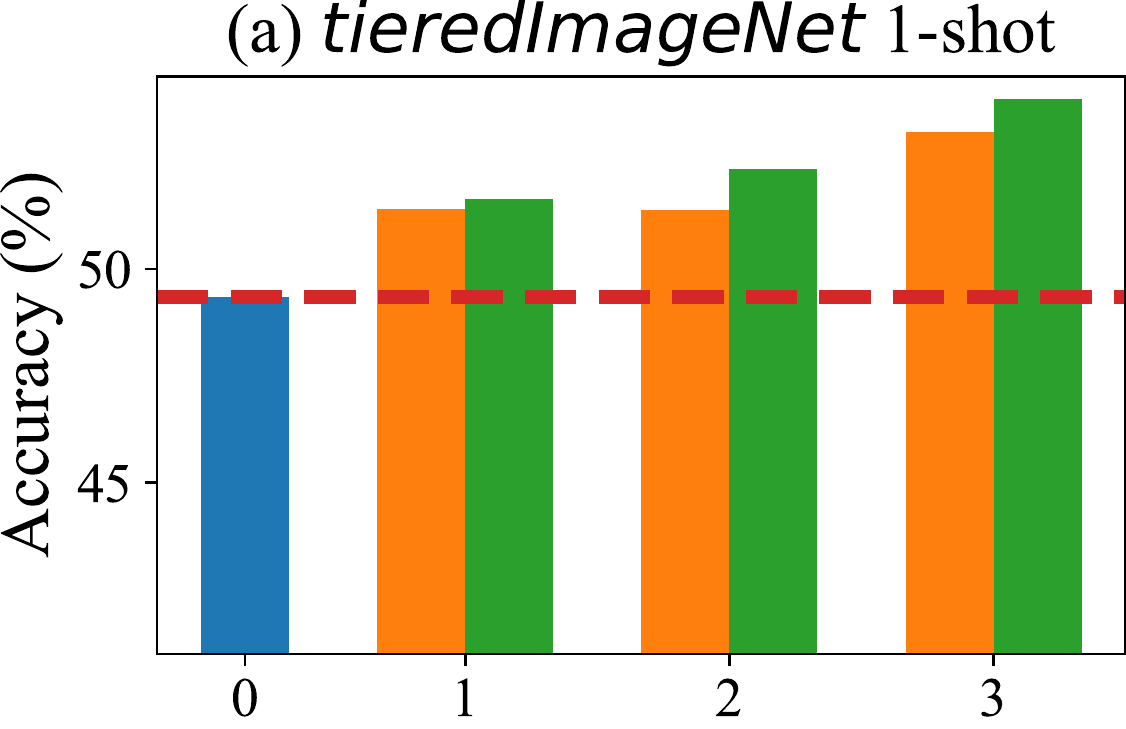}
	\includegraphics[width=0.235\textwidth]{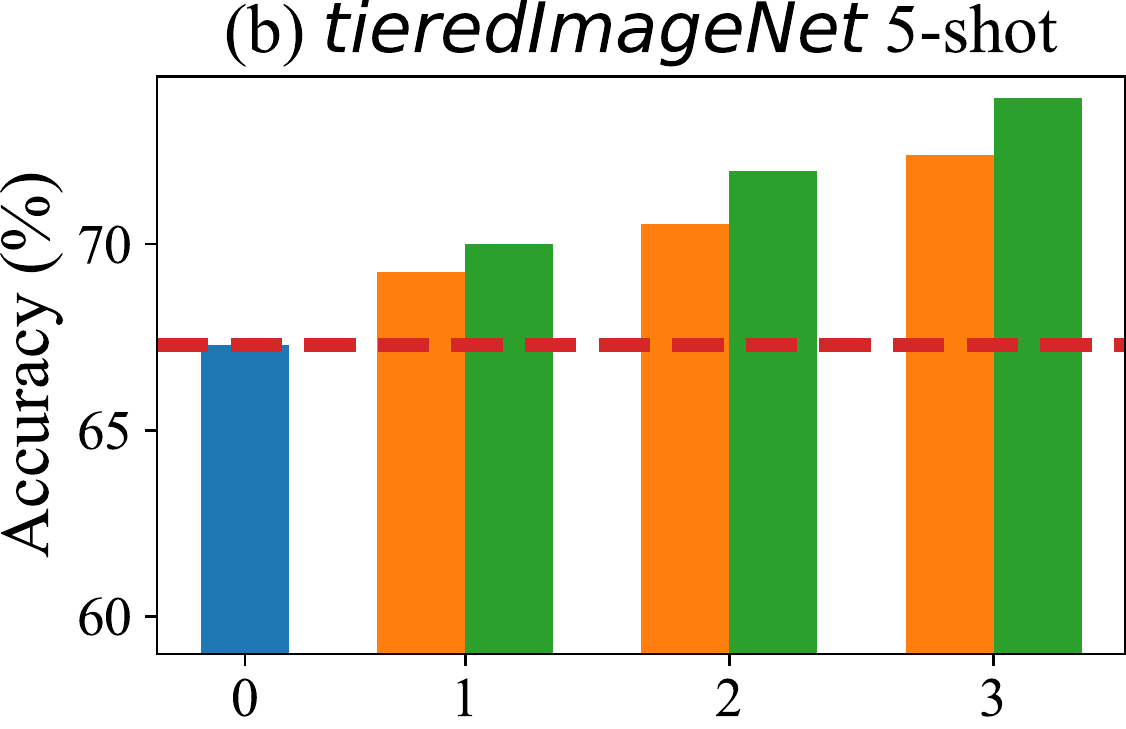}
	\includegraphics[width=0.235\textwidth]{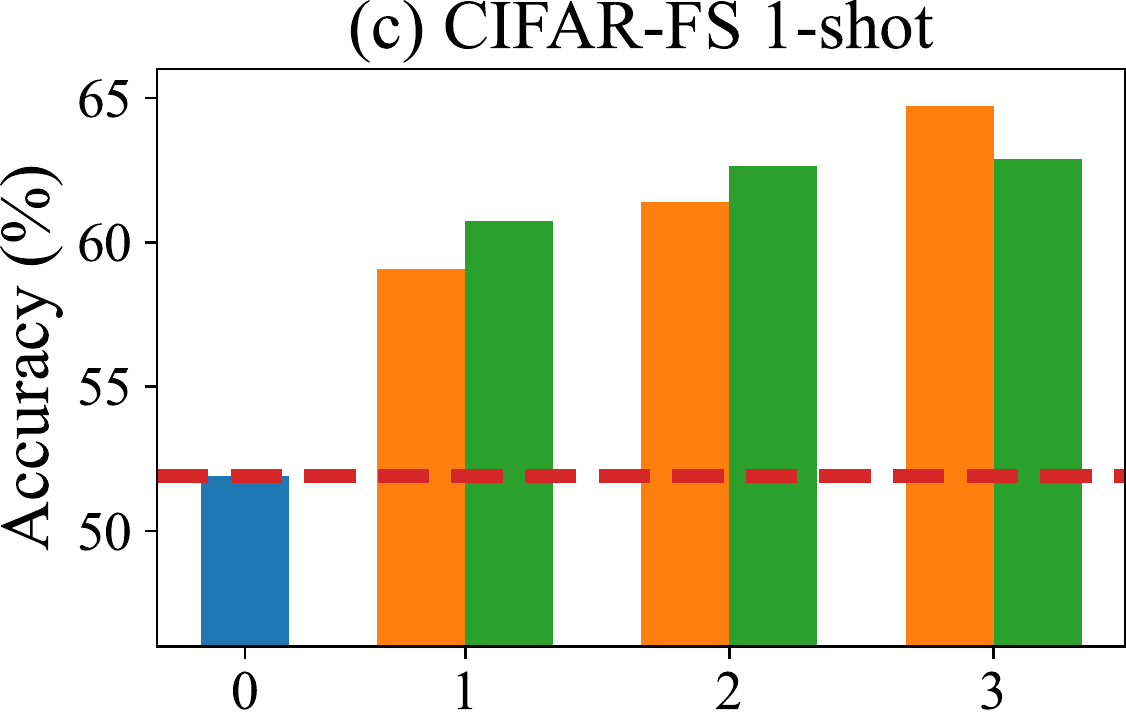}
	\includegraphics[width=0.235\textwidth]{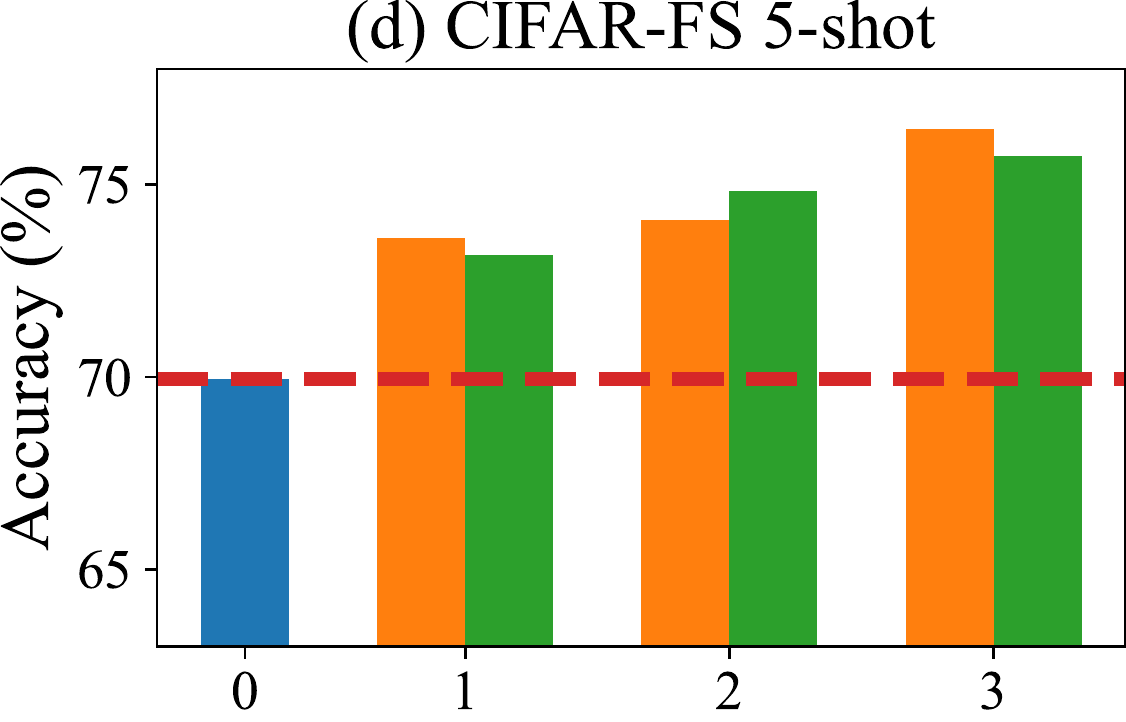} \\
	\includegraphics[width=0.45\textwidth]{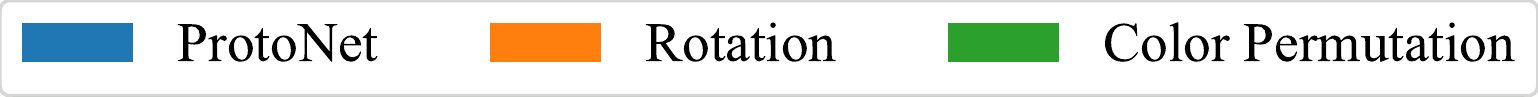}
	\vspace{-5pt}
	\caption{(a)-(d) indiccate different pretext tasks with the same number of child nodes, where (a) and (b) represent \textit{tiered}ImageNet under 5-way 1-shot and 5-shot settings. (c) and (d) represent CIFAR-FS under 5-way 1-shot and 5-shot settings. Red dotted lines represent the performance of baseline (ProtoNet).} 
	\label{fig:dff}
\end{figure}

\noindent \textbf{The effect of different pretext tasks.}
In the main paper, we have reported the results of \textit{mini}ImageNet and CUB-200-2011 with rotation and color permutation tasks under different pretext tasks.
Using the same setting with~\cref{fig:dtt} (a) and (b), we show the results of \textit{tiered}ImageNet and CIFAR-FS datasets in~\cref{fig:dff}.  
From (a)-(d), we find that pretext tasks can bring semantic structure information and improve the generalization ability on all classification datasets.

\clearpage
\bibliographystyle{splncs04}
\bibliography{egbib}

\end{document}